\documentclass[10pt,twocolumn,letterpaper]{article}

\usepackage[pagenumbers]{cvpr} % To force page numbers, e.g. for an arXiv version

\usepackage{graphicx}
\usepackage{amsmath}
\usepackage{amssymb}
\usepackage{booktabs}
\usepackage{multirow}
\usepackage[table]{xcolor}
\usepackage[accsupp]{axessibility}  % Improves PDF readability for those with disabilities.

\usepackage[pagebackref,breaklinks,colorlinks]{hyperref}

\usepackage[capitalize]{cleveref}
\crefname{section}{Sec.}{Secs.}
\Crefname{section}{Section}{Sections}
\Crefname{table}{Table}{Tables}
\crefname{table}{Tab.}{Tabs.}

\newcommand{\loss}{\ensuremath{\mathcal{L}}}
\newcommand*{\affmark}[1][*]{\textsuperscript{#1}}

\def\cvprPaperID{0008} % *** Enter the CVPR Paper ID here
\def\confName{L3D-IVU Workshop}
\def\confYear{2022}

\begin{document}

\title{TDT: Teaching Detectors to Track without Fully Annotated Videos}

\author{
Shuzhi Yu\affmark[1]\thanks{This work was done when Shuzhi Yu was an intern at Google} \;\;\;\;
Guanhang Wu\affmark[2] \;\;\;\;
Chunhui Gu\affmark[2] \;\;\;\;
Mohammed E. Fathy\affmark[2] \\
\affmark[1]Duke University \;\;\;\; \affmark[2] Google LLC \\
{\tt\small shuzhiyu@cs.duke.edu} \;\;\;\; {\tt\small \{guanhangwu,chunhui,msalem\}@google.com}
}
\maketitle

%%%%%%%%% ABSTRACT
\begin{abstract}
% Tracking-by-detection is a popular paradigm to approach the Multi-Object Tracking (MOT) problem, where objects are first detected from each frame and linked together based on the motion and appearance matching. 

Recently, one-stage trackers that use a joint model to predict both detections and appearance embeddings in one forward pass received much attention and achieved state-of-the-art results on the Multi-Object Tracking (MOT) benchmarks. However, their success depends on the availability of videos that are fully annotated with tracking data, which is expensive and hard to obtain. This can limit the model generalization. In comparison, the two-stage approach, which performs detection and embedding separately, is slower but easier to train as their data are easier to annotate. We propose to combine the best of the two worlds through a data distillation approach. Specifically, we use a teacher embedder, trained on Re-ID datasets, to generate pseudo appearance embedding labels for the detection datasets. Then, we use the augmented dataset to train a detector that is also capable of regressing these pseudo-embeddings in a fully-convolutional fashion. Our proposed one-stage solution matches the two-stage counterpart in quality but is 3 times faster. Even though the teacher embedder has not seen any tracking data during training, our proposed tracker achieves competitive performance with some popular trackers (\eg JDE) trained with fully labeled tracking data.

% our proposed tracker outperforms other trackers that do not use annotated tracking data, and stays competitive with some popular trackers (\eg JDE) trained with fully labeled tracking data.

% this simple framework outperforms other methods that do not use full tracking annotations, and stay competitive compared to those that use full annotation.

\end{abstract}

%%%%%%%%% BODY TEXT
\section{Introduction}
\label{sec:intro}
% Story line
% General description of the MOT problem
% The rising one-stage tracker becomes the state of the art: Advantages and disadvantages
% Rethink about the benefits of the traditionally popular two-stage trackers: Advantages and disadvantages
% Weakly supervised scheme: Introduce our weakly supervised scheme: A diagram
% Trying to combine the best of the two worlds
% Benefits of easy training
% Analysis of the system: Better teacher embedders lead to better results, what situation it works
% Contributionhttps://braintex.goog/project/61818eb6c26c06007e99f003#cite.peng2020chained
% Description of the following sections

Tracking people and objects in videos is an important task in computer vision and at the core of many applications. For tracking to be achieved, several sub-tasks need to be solved. The system has to identify (potentially many) objects of interest in the video, \ie \emph{object detection}, and it has to relate the locations of these objects as they move in the video, \ie \emph{data association}. Many approaches have been and continue to be proposed for solving the tracking problem~\cite{SnchezMatilla2016OnlineMT,xiang2015learning,chen2017online,bae2017confidence,zhou2019objects,wang2020JDE,zhang2021fairmot}. One popular approach is tracking by detection. In this approach, detection is made on each video frame and the tracker associates the detections from consecutive frames using various cues such as motion and appearance (a.k.a. appearance embeddings).

Many of the typical tracking algorithms use separately trained and operated components for detection and embedding~\cite{SnchezMatilla2016OnlineMT,xiang2015learning,chen2017online,bae2017confidence}. Recently, one-stage trackers that use a single convolutional neural network (CNN) for joint detection and appearance embedding were introduced~\cite{lu2020retinatrack,wang2020JDE,zhang2021fairmot}. As the detections and their embeddings are generated in one forward pass, these approaches are relatively faster than traditional two-stage ones. In addition, they achieve state-of-the-art performance on many Multi-Object Tracking (MOT) benchmarks~\cite{wang2020JDE,zhang2021fairmot}.

\begin{figure}[t]
    \centering
    \includegraphics[width=0.98\linewidth]{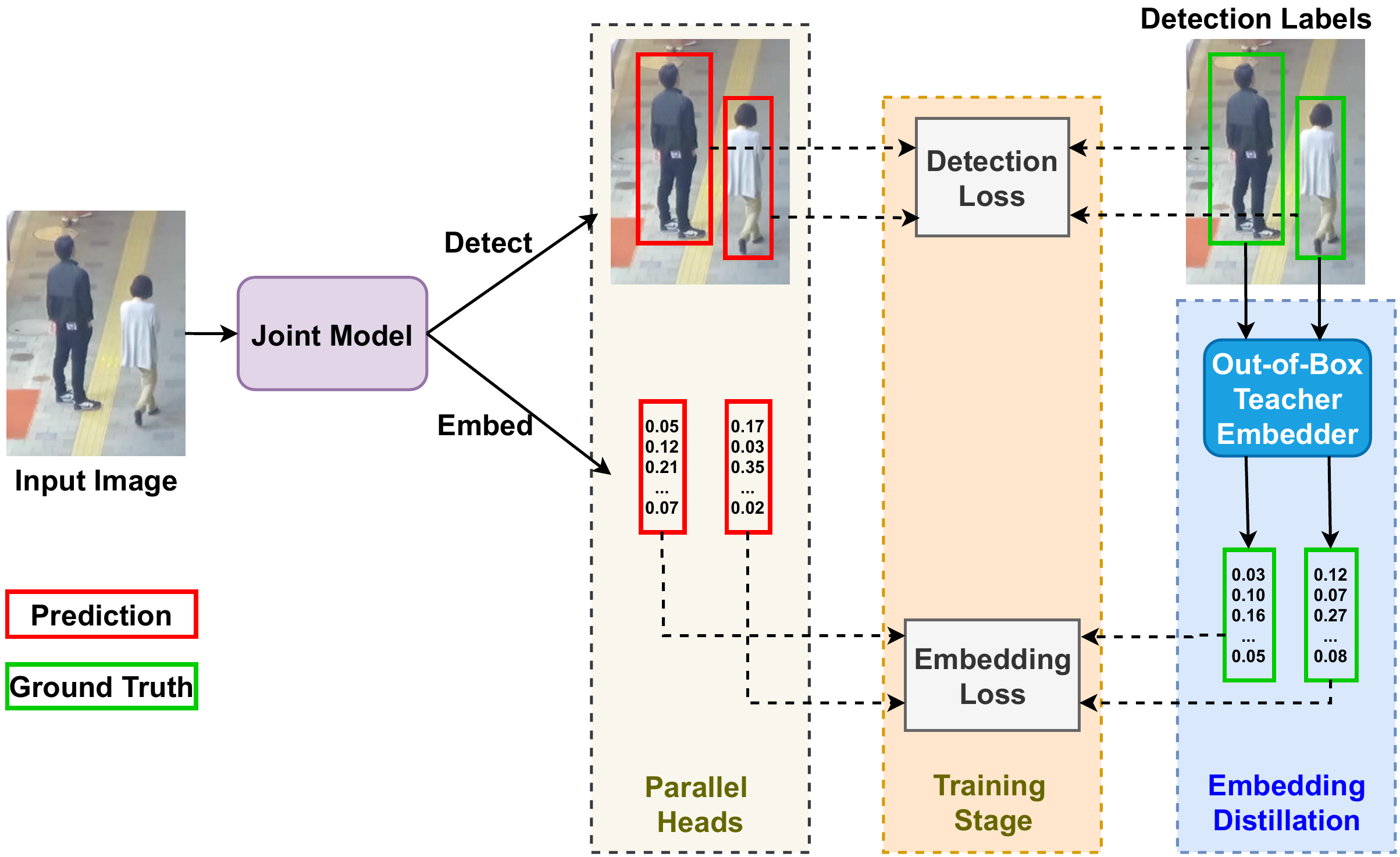}
    \caption{The training stage of our proposed one-stage tracking framework using only detection boxes as labels. We augment a detection example (top right image) with pseudo ground-truth embeddings (shown as the embedding vector in green) through a data distillation approach. These teacher embeddings are obtained by feeding the cropped image patches (green bounding boxes) into a teacher embedder (a pre-trained out-of-box Re-ID network). A joint model takes an input image, and jointly detects objects of interest (red bounding boxes) and generates an embedding (red embedding) for each object. The augmented dataset is used to supervise both predictions. This system is so flexible that different detectors and teacher embedders can be used.}
    \label{fig:wsjm_demo}
\end{figure}

However, the success of these joint models depends heavily on the availability of fully annotated tracking data including both detection and identification labels. They need both labels to simultaneously supervise the detection and embedding components. Such datasets are more restricted in nature and quantity than typical detection-only and identity-only datasets as they are much harder and more expensive to annotate. The lack of fully-annotated videos from diverse conditions can negatively impact the ability of one-stage trackers to generalize. In particular, they work very well on surveillance videos (\ie the same data that is used to train them) but fail on videos from media and other domains where large people are likely to appear.

Reconsidering the two-stage models, although they are relatively slower, they do not require the expensive tracking labels for training. The two components, namely detector and embedder, are trained separately on their specialized datasets~\cite{yu2016poi}. Such decoupling makes it possible to use less restrictive and more diverse detection and identification training sets, leading to better generalization of the individual components and the overall tracking system.

% Our joint model can be integrated into any data association system as other joint models.
Accordingly, we propose in this paper a practical and effective approach (\cref{fig:wsjm_demo}) that combines the speed of joint models with the lower training cost and better generalization of two-stage ones. In particular, we introduce a weak supervision framework that can train a joint model without fully annotated videos. The framework follows the teacher-student learning scheme~\cite{Hinton2015DistillingTK,Romero15FitNets} for training the embedders, \ie embedding distillation. This is done by augmenting a given detection dataset with pseudo ground-truth embeddings generated by a Re-Identification (Re-ID) model independently trained on Re-ID datasets~\cite{zhou2019osnet,zhou2021osnet}. During training, the detector training loss function is augmented with an embedding distillation loss term to make sure the detection and embedding heads of the joint model are adequately supervised. This way, any detection dataset can be converted to a detection and embedding distillation one that can be used to train a joint model. Our experimental results suggest that training our model on the additional augmented detection dataset improves the overall detection and tracking performance (see ~\cref{fig:dataset_impact}). Furthermore, we show empirically that the overall tracking performance is enhanced with better teacher embedders (\cref{tab:teacher_embedders}). 

% Our system works especially well on scenarios where pedestrians are large and limited overlapping, which is closer to the Re-ID datasets the teacher embedder is trained on. This indicates many potentials of the proposed weakly supervised scheme as better Re-ID networks are being proposed and trained on more diverse Re-ID datasets.

% In addition, the resulting model potentially achieves the same performance and generalization of the corresponding two-stage model but at a faster speed.

In addition, the training of the embedding is simplified in our weakly supervised framework. In general, there are two losses on embedding, namely the triplet loss~\cite{weinberger2009distance,schroff2015facenet,voigtlaender2019mots} and the cross-entropy loss. The implementation of the triplet loss can be tricky since effectively selecting such triplets from a large sampling space is a non-trivial problem~\cite{schroff2015facenet}. The cross-entropy loss models each person in the training videos as a class, which can make training harder and slower as we increase the number of labeled people in the training set. In comparison, our weakly supervised loss is straightforward and easy to implement. A simple least squares error between the predicted and pseudo ground truth embedding is shown to work well, without any sophisticated weighting algorithm to balance the detection and embedding losses.

% A triplet loss measures the sum of some similarity between the embeddings of the anchor and negative samples and some dissimilarity between the anchor and the negative sample. The cross entropy loss models each person in the training videos as a class, and the embedder head predicts $N$ scores for $N$ unique identities in the videos on each detected bounding box. However, there is a limit on the number of identities the loss can handle and it constraints the overall computational power of the model since the embedder head can be memory heavy under large $N$. 

% The entropy loss may lead to more discriminative features for the tracking application and is more stable~\cite{wang2020JDE}. 

% Describes alternatives to our method and their disadvantages.
Another potential way of avoiding fully annotated tracking data is to alternatively train the detection and embedding heads of the joint model on detection and Re-ID datasets, respectively. Compared to our approach, this leads to a much more complicated training process involving more hyperparameters (\eg learning rates, data sampling schemes, training schedules, \etc for different heads). In particular, it is not clear how to pick generally good settings that would work well for different combinations of datasets. Our method avoids these issues by using distillation to generate a single dataset, instead of two radically different ones.

% In addition, alternating the training of the two heads generate different gradients for updating the backbone features, which may lead to longer convergence time. , for training the two heads simultaneously using an approach akin to typical detector training.

% Interestingly, our joint model outperforms a state-of-the-art joint model, FairMOT~\cite{zhang2021fairmot}, trained by the fully annotated tracking labels, on such scenarios~\cref{tab:compare_det_emb}. On the other hand, our joint model has difficulties on tracking small pedestrians and those with large overlaps or occlusions since those samples are not common in the Re-ID datasets. However, these cases are common in the tracking benchmarks. Thus, FairMOT works better on these cases since their embedders are trained on these tracking data. 

% One reason can be that the surveillance videos are more common in the tracking datasets.

To our best knowledge, this is the first work that trains joint tracking models without using any fully annotated tracking data. Furthermore, our tracker, named TDT-tracker (\ie Teaching Detector to Track), runs 3 times faster than the counterpart two-stage model and stays competitive on the benchmark MOT challenges.

%  Such systems are flexible in terms of the training labels, which potentially leads to better generalization ability. The loss on the embedder head is much easier and straightforward to implement. The framework can be applied into any object detector, and we hope this paper facilitate the research direction of weakly supervised tracker system. 

In the rest of the paper, \cref{sec:rel_work} discusses the related work, \cref{sec:method} explains the proposed approach, \cref{sec:exp_res} shows the empirical evaluation of our model, and \cref{sec:disc_con} ends the paper with a conclusion.

%------------------------------------------------------------------------
\section{Related work}
\label{sec:rel_work}
\paragraph{Two-stage tracking system:}
Many methods took public detections~\cite{felzenszwalb2009object,yang2016exploit,dollar2014fast} as input and focused on improving the data association step~\cite{SnchezMatilla2016OnlineMT,xiang2015learning}. They used historical states in the tracklet or hand-crafted features as cues to connect objects across frames. Some approaches~\cite{chen2017online,bae2017confidence,yu2016poi,Mahmoudi2018MultitargetTU,wojke2017simple} used CNNs to extract embeddings as a more robust representation of the objects. The detections and embeddings are produced sequentially, and thus, such systems are referred as the two-stage tracking system. Xiao~\etal \cite{xiao2017joint} proposed to use the low-level features more efficiently by sharing them between the detection and embedding branches. However, the speed is not close to the real time since the two components are essentially still sequential.

% This is similar to what Re-ID community does since a discriminative embedding is essential to solving the task.
\paragraph{One-stage tracking system:}
% Inspired by the success of multi-task deep learning in computer vision~\cite{zhang2014facial,dai2016instance,Liu2019endtoend}, 
%  Fang~\etal used a Recurrent Autoregressive Network (RAN) to predict probable locations of the tracking targets based on the trajectory history~\cite{fang2018recurrent}, which helps in the event of occlusion.

A recent approach, named Joint Detection and Embedding (JDE), achieved almost real-time MOT by jointly predicting bounding boxes and their associated embeddings using the same CNN in a single forward pass~\cite{wang2020JDE}. A subsequent work, FairMOT~\cite{zhang2021fairmot} that is built on the CenterNet~\cite{zhou2019objects}, further improved the performance on the benchmark tracking datasets by carefully balancing the two tasks during training. The joint learning scheme was further extended to include the data association step as well and achieved promising results~\cite{pang2020tubetk,fang2018recurrent,peng2020chained,wang2021multiple,Wang2021joint}. Particularly, TubeTK~\cite{pang2020tubetk} jointly modeled the spatial and temporal directions under a 3D CNN framework. Chained-Tracker chains paired objects from adjacent frames to directly integrate the data association step into the end-to-end training~\cite{peng2020chained}. Wang~\etal proposed to add a learnable correlation module to the joint model~\cite{wang2021multiple} and achieved even better results. All of these trackers are trained on the tracking annotations~\cite{MOTChallenge2015,MOT16,MOTChallenge20,dollar2009pedestrian,xiao2017joint,Zheng2017PersonRI}.

In comparison, our proposed training framework does not require expensive fully annotated tracking data. Instead, we combine a detector and an embedder through a data distillation approach~\cite{radosavovic2018data}. We also borrow ideas from knowledge distillation, which aims to compress a complicated model to a simpler student model~\cite{Hinton2015DistillingTK,Romero15FitNets}. We empirically found that an embedder head consisting of merely two convolutional layers is able to mimic the embedding of a much larger teacher embedder given the backbone features, and thus, it compresses the teacher embedder model.  

Not much work focuses on mitigating the scarcity of the fully annotated tracking data on the MOT problem. Fabbri~\etal created a synthetic dataset named MOTSynth that contains ground-truth labels for detection and tracking~\cite{fabbri21motsynth}, and empirically showed that the tracker can achieve promising results with training only on this dataset. However, whether these sythetic datasets generalize to real world is still an open question. A concurrent work, KDMOT~\cite{zhang2021boosting}, explores in a similar direction with different purposes. The method still pre-trains a joint model on some fully-annotated tracking data before using a teacher model to supervise its embedding head. In comparison, our goal is to understand how well such a joint model can perform without any limited tracking data. Thus, our model does not see any tracking label during training. In addition, our teacher embedder is only trained on one Re-ID dataset. 

% Tracktor uses the bounding box regression of the Faster R-CNN~\cite{ren2015faster} to regress the previous bounding box to the new location~\cite{Bergmann2019ICCV} without using any tracking labels. However, the tracking performance largely increased with Re-ID features trained on the tracking data. FairMOT~\cite{zhang2021fairmot} pre-trained their joint model on a detection dataset augmented by some artificially created ID labels.

% Recently, a new problem of joint learning MOT and segmentation (MOTS) has drawn some attention since the two problems are highly related~\cite{voigtlaender2019mots,xu2020segment}. Datasets with both tracking and segmentation annotation are even rarer, and our distillation framework can be easily extended to weakly supervise the segmentation head by a segmentation teacher.

\paragraph{Object detector and Re-ID networks:}
% The community of the object detection problem has been actively proposing detectors of different object types with increasing accuracy and faster inference speed

We need to choose a suitable detector and appropriate teacher embedder for our proposed framework. In principle, most of the deep object detectors~\cite{girshick2015fast,ren2015faster,redmon2016you,zhou2019objects,law2018cornernet,liu2016ssd} and any embedding-based Re-ID network~\cite{li2014deepreid,ahmed2015improved,varior2016gated,sun2018beyond} can be used in our system. In our experiment, we choose the one-stage detector (\ie without Region-of-Interest pooling) RetinaNet~\cite{lin2017focal} as our backbone detector given its simple design. Since people appearing in videos can come in various sizes and different imaging conditions, we prefer to use Re-ID methods that cover multi-scale features and generalize well across domains~\cite{chang2018multi,liu2017hydraplus,wang2018resource,zhou2019osnet}. Among the many good choices, we choose OSNet~\cite{zhou2019osnet} as our teacher embedder. Although it is not the state-of-the-art embedder on the Re-ID benchmarks anymore, its simplicity, lightweight design, and well-maintained code base~\cite{zhou2021osnet} make it a good candidate to demonstrate the effectiveness of our method.

% S{\'a}nchez-Matilla\etal proposed an data association procedure that use detections of low and high confidence in a different way~\cite{SnchezMatilla2016OnlineMT}. Xiang\etal models the similarity learning in the data association as learning a policy for the Markov Decision Process. Features are based on the historical states of the tracklet~\cite{xiang2015learning}. Chen \etal proposed to utilize the multi-level features from the backbone CNN for both instance classification (between identities) and category classification (between different types of objects)~\cite{chen2017online}. Bae and Yoon proposed a data association method based on tracklet confidence score and also a deep appearance learning network (fined-tuned on the tracking dataset)~\cite{bae2017confidence}.

%------------------------------------------------------------------------
\section{Method}
\label{sec:method}
% Problem formulation: Input, output, notation
% Architecture: Joint model (Backbone, multiple features, detector head, embedding head)
% Overall pipeline, By a Re-ID teacher embedder, generally we can use any teacher embedder, briefly explain the one we used, explain how embedding is cropped
% Losses: Detection related loss, embedding loss, weighted
% Data association

\setlength{\belowcaptionskip}{-6pt}
\begin{figure}[t]
    \centering
    \includegraphics[width=0.98\linewidth]{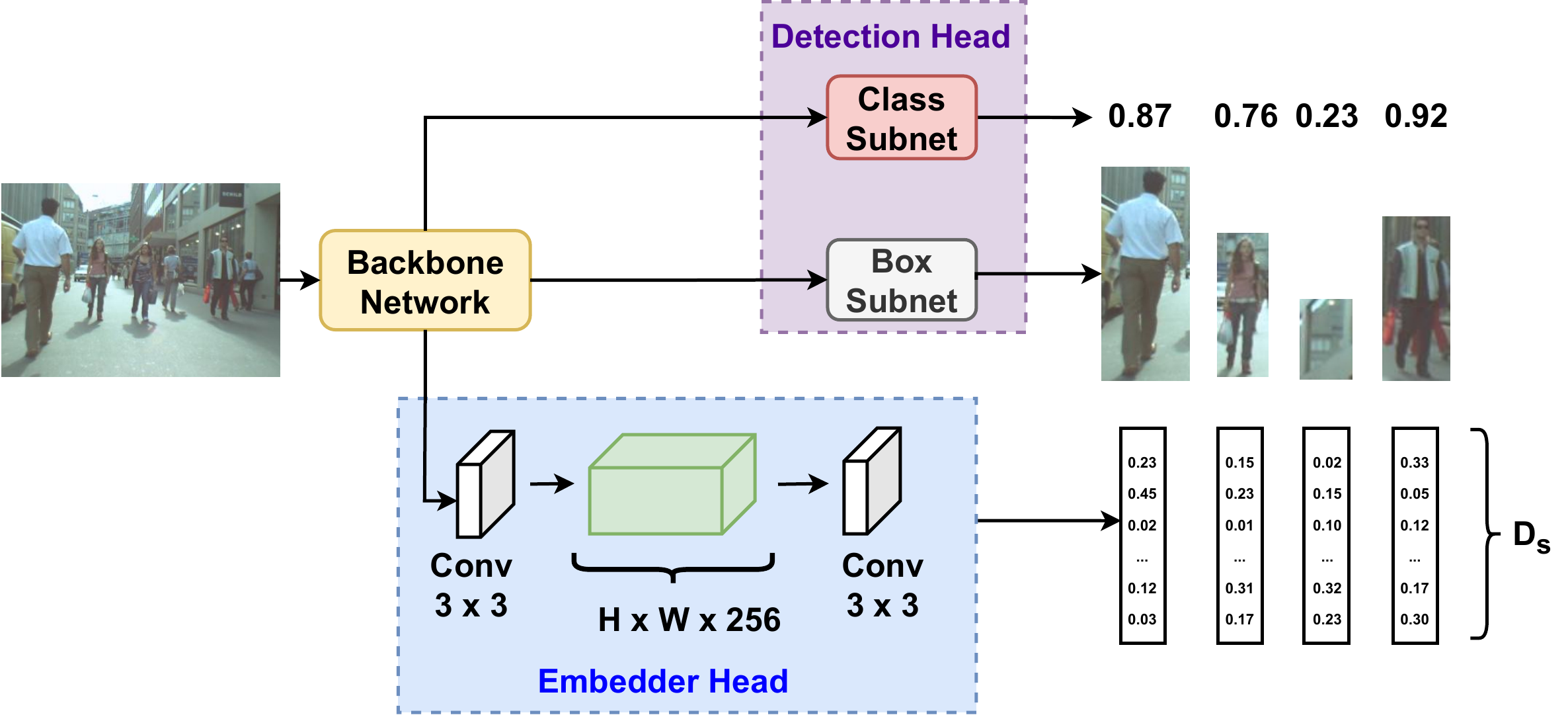}
    \caption{Our proposed joint model, adapted from RetinaNet~\cite{lin2017focal}. The newly added embedder head, consisting of merely two convolutional layers, is parallel to the detection head. Given an input image, the backbone network extract multi-level features, which are fed into the three sub-networks simultaneously. The box sub-net predicts the location of the bounding boxes containing the objects of interest. The class sub-net computes a confidence score of every class for each predicted bounding box. The embedder head generates an embedding for each bounding box, representing the discriminative features of that object.}
    \label{fig:arch_demo}
\end{figure}

\subsection{Problem formulation}
Given a detection dataset and a pre-trained (teacher) Re-ID network, the goal is to train a one-stage tracker that achieves good performance on a target tracking dataset. The detection dataset only provides the ground-truth bounding boxes $B_i\in \mathbb{R}^{K\times 4}$, where $K$ is the number of ground-truth bounding boxes in the $i^{th}$ image. We do not use any identification labels of any tracking dataset in our network training.

% The identification labels of any tracking datasets cannot be used in training either the Re-ID network or the tracker.

% This requires that the joint model can accurately localize the multiple objects in each frame and generate discriminative embedded features for each detected object. 

\subsection{Model architecture}
% The features of finer levels contains the information of the features in the coarser levels through skip connections. This is useful to pedestrian detection since a video usually contains people of different sizes.

We choose RetinaNet~\cite{lin2017focal} as our base detector, and add a simple embedder head to the RetinaNet, making it a joint model of detection and embedding. RetinaNet can be combined with different available architectures, e.g. MobileNet~\cite{Howard17mobilenets}, ResNet~\cite{he2016deep}, and SpineNet~\cite{du2020spinenet} to predict detections. Specifically, RetinaNet uses a Feature Pyramid Network (FPN)~\cite{lin2017feature} to generate multi-level features as well as a classification subnet and a box regression one that utilize these multi-level features to make predictions on each of the FPN levels.

As shown in \cref{fig:arch_demo}, an embedder head (highlighted in the blue rectangle) is added to RetinaNet, parallel to its detection head. The embedder branch is a cascade of 2 convolutional layers with a batch normalization layer~\cite{ioffe2015batch} in between. This head predicts an embedding for each anchor box at each level. This means that the output embeddings for level $l$ have size $D_s\times S\times R\times H_l\times H_W$, where $S$ and $R$ are the number of scales and aspect ratios in the anchors respectively, $H_l$ and $H_w$ are the height and width of level $l$ in the feature pyramid respectively, and $D_s$ is the dimension of the output embedding (\ie student embedding).

At inference time, we apply non-maximum suppression (NMS) at threshold 0.5 to reduce the number of anchors. We only take the top 100 or less anchors, and remove those anchors with classification scores smaller than 0.05. The predicted embedding is normalized before being used in the data association process.

\subsection{Weak supervision framework}
\label{sec:weaksup_framework}
\Cref{fig:wsjm_demo} shows the proposed weak supervision framework. We generate a pseudo ground-truth embedding, \ie the teacher embedding, for each ground-truth bounding box provided by the detection dataset. The areas corresponding to each bounding box are cropped independently and resized to the resolution expected by the teacher embedder, say $256\times 128$ as in our experiments~\cite{zhou2019osnet}. These image patches are fed into the teacher embedder, and a teacher embedding $f\in\mathbb{R}^{D_t}$ of dimension $D_t$ (\ie dimension of the teacher embedding) is generated and attached to each of these bounding boxes. The student embedding can have the same or less number of dimensions than the teacher embedding, that is $D_s \leq D_t$. 

We create these pseudo embedding labels for each detection dataset beforehand and store them for the training process. Therefore, this is a one-time computational cost. The memory overhead depends on the density of the bounding boxes in the detection dataset and the number of dimensions of the teacher embeddings. For example, we use a teacher embedding of 512 dimensions~\cite{zhou2019osnet,zhou2021osnet}, and the storage overhead of the detection datasets used in our experiments ranges from $9.3\%$ to $73.3\%$ of the original storage.

Following RetinaNet, each anchor is at most assigned to one ground-truth bounding box based on the intersection-over-union (IoU). If an anchor is assigned to a ground-truth bounding box, it is also assigned the teacher embedding of that bounding box. There is no pseudo ground-truth embedding for those unassigned anchors.

\paragraph{Teacher embedder}
% Describe the teacher embedder we chose to use.
We choose a pre-trained out-of-box Re-ID network as our teacher embedder. The data association step in the tracking process requires the embeddings to be discriminative so that the same object occurrences can be associated across different frames. This is also required in the person re-identification task and thus, the Re-ID network is a good candidate as our teacher embedder. We choose OSNet~\cite{zhou2019osnet,zhou2021osnet} as the teacher embedder in our proposed teacher-student framework due to its multi-scale design and well-maintained lightweight implementation~\cite{torchreid}. 

% OSNet explicitly extracts features of multiple scales and dynamically fuse them by a carefully designed gating mechanism. 

% Different scales of features are embedded by the stack of different number of residual blocks of the same architecture. The gating method assigns different weights, computed based on the input image, to different scales of features in the fusion stage. The light weight attributes to the usage of depthwise separable convolutions~\cite{Howard17mobilenets,chollet2017xception}. We refer readers to read OSNet paper for more details.

\subsection{Losses}
\label{sec:losses}
We use different types of losses for these three types of predictions, namely classification, bounding box localization, and embeddings. These losses are measured on all the valid anchor boxes across all the levels defined by the FPN.

Specifically, we use focal loss~\cite{lin2017focal} $\loss_c$ as the supervision signal on the predicted classification score by the classification subnet to mitigate the class imbalance problem. Although there may be many people in a video, the class imbalance issue still exists due to the large number of anchors. The focal loss reduces the contribution of those easy samples to the final loss so that the network can learn on the hard samples. 

% The total focal loss $\loss_c$ of an image is the focal loss over all anchors divided by the number of anchors assigned to the objects of interest~\cite{lin2017focal}, \eg people.

Huber loss~\cite{hastie01statistical} $\loss_b$ is used on the box regression because of its robustness to outliers. We use the least square loss (L2 loss) as the loss measured between the teacher embeddings $f$ and the predicted student embeddings $\hat{f}$, that is, 
\begin{equation}
    \loss_e(f, \hat{f}) = \sum_{i=1}^{D_s}(f_i - \hat{f}_i)^2\;,
\end{equation}
where $f_i$ is the $i^{th}$ element of the embedding. If $D_s < D_t$, we use the first $D_s$ elements in the teacher embedding $f$ as the supervision signal. Recall that $D_t$ is the dimension of the teacher embedding. The re-normalization of the new teacher embedding is not required during training. The embedding loss $\loss_e$ of an image is averaged embedding loss over all the assigned anchors.

The total loss is weighted over the three losses,
\begin{equation}
    \loss = \alpha_c \loss_c + \alpha_b \loss_b + \alpha_e \loss_e\;.
\end{equation}
We find that it is robust to a range of different weighting schemes.

\subsection{Data association}
Our joint model can be easily integrated into different existing data association algorithms. Particularly, we use the association system proposed by FairMOT~\cite{zhang2021fairmot}. A tracklet represents the current state of the corresponding track. The first batch of tracklets are initialized based on the detections in the first frame. The Kalman filter~\cite{kalman1960new} is used to predict a likely location of the tracklet in the next frame. Both the motion and the embedding of the objects are used to match the detections in the new frame and the existing tracklets in the tracklet pool, by the Hungarian algorithm~\cite{kuhn1955hungarian}. We found that the threshold used in the matching distance can make a difference, since our embedding space is likely different from theirs. The unmatched tracklets and detections are further matched by thresholding the IoU at 0.5.

%------------------------------------------------------------------------

\section{Experimental results}
\label{sec:exp_res}
% The embedding loss seems not to affect box and classification loss (We find alpha_e is quite robust).

\subsection{Datasets and metrics}
Following JDE \cite{wang2020JDE} and FairMOT \cite{zhang2021fairmot}, we integrate several datasets as a large training dataset to avoid training bias. Particularly, we use ETH dataset~\cite{ess2008mobile}, CityPersons dataset~\cite{zhang2017citypersons}, and CrowdHuman dataset \cite{shao2018crowdhuman}, which only provide detection labels. In addition, we use fully annotated tracking datasets CalTech dataset~\cite{dollar2009pedestrian}, MOT17 dataset~\cite{MOT16}, CUHK-SYSU dataset~\cite{xiao2017joint}, and PRW dataset~\cite{Zheng2017PersonRI}. However, we only use their detection labels for training our joint models and ignore the ID labels. The embedding branch is supervised by the teacher embedders that are not trained on these tracking datasets either.

The teacher embedders used in our experiments are only trained on the MSMT17 dataset~\cite{wei2018person}, which consists of 180 hours of videos and identity labels for person re-identification task. We use the teacher embedders to augment the mentioned detection datasets with pseudo embedding labels, following the procedure described in \cref{sec:weaksup_framework}.

We use \textit{BaseTrainSet} to denote the collection of all these augmented detection datasets excluding CrowdHuman. BaseTrainSet is used to train various models in our ablation studies (if not explicitly mentioned) and in \cref{tab:1vs2}. We use six video sequences from the MOT15~\cite{MOTChallenge2015} training set as the validation dataset in the ablation studies and in \cref{tab:1vs2}, which do not include the overlapped sequences with the MOT17 dataset. We also evaluate our model, trained on both the BaseTrainSet and augmented CrowdHuman datasets, on the private MOT16 and MOT17 benchmark~\cite{MOTChallenge2015,MOT16} to compare with other work (\cref{tab:compete_stoa}). Our model is further fine-tuned on the MOT20~\cite{MOTChallenge20} training set before evaluated on the private MOT20 benchmark (\cref{tab:compete_stoa}).

We evaluate both the detection and tracking abilities of our models. We use Average Precision (AP) at IoU 0.5 to measure the performance of detection~\cite{wang2020JDE,zhang2021fairmot}. Tracking performance is measured by the popular CLEAR metrics~\cite{bernardin2008evaluating} and IDF1 score~\cite{ristani2016performance}.

\subsection{Implementation details}
% Hyber parameters of the loss, loss weights.
We use most of the default settings of RetinaNet~\cite{lin2017focal} in our base detector. Particularly, the FPN consists of features from level 3 to 7. We use three scales and three aspect ratios (0.25, 0.5, and 1.0) of anchors. Both the class subnet and box subnet consist of four convolutional layers of feature size 128. The embedder head consists of two convolutional layers of filter size $3\times 3$. The intermediate feature size is 256. We experiment on the output embedding size of 64, 128, 256, and 512 in an ablation study (\cref{tab:dim_analysis}), and use 128, combined with a ResNet-34 backbone architecture, in the benchmark evaluation (\cref{tab:compete_stoa}). The teacher embedder~\cite{zhou2019osnet,zhou2021osnet} generates embeddings of size 512.

The input image size is $608\times 1088$ for ablation studies and \cref{tab:1vs2}, and $1080\times 1920$ for private benchmark evaluation (\cref{tab:compete_stoa}). Since our teacher embedder is not trained on the similar tracking datasets, we found that clearer human figures resulting from the larger resolution lead to better tracking performance. We use standard data augmentation techniques including random horizontal flipping and random scaling with scaling ratio from 0.5 to 2.0.

For the weighted loss, we set $\alpha_c = 1$ for the focal loss on classification, $\alpha_b = 50$ for the Huber loss on the bounding box regression, and $\alpha_e = 10$ for the L2 loss on embedding. We found that both the detection and tracking performance are robust to the weight of the embedding loss in a range of 2 to 10. We set $\alpha$ as 0.25 and $\gamma$ as 1.5 in the focal loss; the $\delta$ in the Huber loss is 0.1.

We set initial learning rate as 0.15 and use the cosine decay schedule~\cite{Loshchilov17SGDR}. The warm up learning rate is 0.001 for the first 2000 iterations. We use batch size of 16. The models are trained with 300K to 500K iterations depending on the settings. All inferences are computed on a single Nvidia P100 GPU.

\subsection{Main results}
% In this section, we compare our joint model with the counter-part two-stage model on the validation dataset (\cref{tab:1vs2}), and with other existing trackers on the benchmark datasets to show our advantage of speed and tracking performance (\cref{tab:compete_stoa}).

\subsubsection{TDT-tracker versus two-stage tracker}
\Cref{tab:1vs2} compares our TDT-tracker and its counterpart two-stage tracker. The detectors of both trackers have the same architecture and are trained under the same training settings. From \cref{tab:1vs2}, we see that the joint learning has a negative effect on the detection performance with 1.2\% (0.924 to 0.913) decrease on the Average Precision (AP) of detection. However, the one-stage tracker achieves much better tracking performance with much faster speed. For example, the one-stage tracker achieves 3.3\% better MOTA (0.760 to 0.785) and 37.4\% (91 to 57) less ID switches with 3.2 times faster speed (3.03 to 9.61 frame per second).

\setlength{\belowcaptionskip}{-6pt}
\begin{table}[h]
  \centering
  \begin{tabular}{@{}c|ccccc@{}}
    \toprule
    Tracker & MOTA $\uparrow$ & IDF1 $\uparrow$ & IDs $\downarrow$ & AP $\uparrow$ & FPS $\uparrow$ \\
    \midrule
    Two-stage & 0.760 & 0.769 & 91 & \textbf{0.924} & 3.03 \\
    TDT-tracker & \textbf{0.785} & \textbf{0.787} & \textbf{57} & 0.913 & \textbf{9.61} \\
    \bottomrule
  \end{tabular}
  \caption{Comparison between the two-stage model and our joint model (TDT-tracker). Both models use the same detector architecture, and the joint model has an additional embedder head (\cref{fig:arch_demo}). We use an OSNet~\cite{zhou2019osnet,zhou2021osnet} as the embedder in the two-stage tracker as well as the teacher embedder for training our TDT-tracker.}
  \label{tab:1vs2}
\end{table}

\begin{table*}[h]
  \centering
  \begin{tabular}{@{}l|l|ccccc|c@{}}
    \toprule
    Dataset & Tracker & MOTA $\uparrow$ & IDF1 $\uparrow$ & MT $\uparrow$ & ML $\downarrow$ & IDs $\downarrow$ & Use Tracking Labels? \\
    \midrule
    % \multirow{4}{*}{MOT15} & \multirow{3}{*}{Yes} & RAN~\cite{fang2018recurrent} & 56.5 & 61.3 & 45.1\% & 14.6\% & \textbf{428} \\
    % & & TubeTK~\cite{pang2020tubetk} & 58.4 & 53.1 & $39.3\%$ & $18\%$ & 854 \\
    % & & FMOT~\cite{zhang2021fairmot} & \textbf{60.6} & \textbf{64.7} & \textbf{47.6\%} & \textbf{11.0\%} & 591 \\
    % \cline{2-8}
    % % & \multirow{1}{*}{No}& TDT-RN34 & 52.06 & 58.26 & $31.76\%$ & $15.40\%$ & 628 & \\
    % & \multirow{1}{*}{No}& TDT & 51.82	 & 62.32 & 44.24\% & 12.34\% & 642 \\
    % \midrule
    
    %%%%%%%%%%%%%%%%%%%%%%%%%%%%%%%%%%%%%%%%%%%%%%%%%%%%%%%%%%%%%%%%%%%%%%%%%%%%%%%%%%%%%%%%%%%%%%%%
    \multirow{8}{*}{MOT16} & JDE$^*$~\cite{wang2020JDE} & 64.4 & 55.8 & $35.4\%$ & $20.0\%$ & 1544 & \multirow{3}{*}{Yes} \\
    & ChainedTrackerV1$^*$~\cite{peng2020chained} & 67.6 & 57.2 & 32.9\% & 23.1\% & 1897 & \\
    & FairMOT$^{*}$~\cite{zhang2021fairmot} & \textbf{74.9} & \textbf{72.8} & \textbf{44.7\%} & \textbf{15.9\%} & \textbf{1074} & \\
    \cline{2-8}
    %%%%%%%%%%%%%%%%%%%%%%%%%%%%%%%%%%%%%%%%%%%%%%%%%%%%%%%%%%%%%%%%%%%%%%%%%%%%%%%%%%%%%%%%%%%%%%%%
     &  SORTwHPD16~\cite{bewley2016simple,zhang2021fairmot} &  59.8 &  53.8 & 25.4\% & 22.7\% & 1423 & \multirow{5}{*}{No}\\
     & DeepSORT\_2~\cite{wojke2017simple} & 61.4 & 62.2 & 32.8\% & 18.2\% & \textbf{781} &\\   
     & CNNMTT~\cite{Mahmoudi2018MultitargetTU} & 65.2 & 62.2 & 32.4\% & 21.3\% & 946 & \\
     & POI~\cite{yu2016poi} & \textbf{66.1} & \textbf{65.1} & 34.0\% & 20.8\% & 805 &  \\
     & TDT-tracker$^*$ (Ours) & 64.1 & 61.5 & \textbf{36.5\%} & \textbf{16.5\%} & 1391 & \\
    % & TDT-tracker$^{*}$ (Ours) & 63.2 & 64.5 & 29.4\% & 23.7\% & 835 &  \\
    \midrule
    %%%%%%%%%%%%%%%%%%%%%%%%%%%%%%%%%%%%%%%%%%%%%%%%%%%%%%%%%%%%%%%%%%%%%%%%%%%%%%%%%%%%%%%%%%%%%%%%
    \multirow{6}{*}{MOT17} & CenterTrack$^{*}$~\cite{zhou2020tracking} & 67.8 & 64.7 & $34.6\%$ & $24.6\%$ & \textbf{2583} & \multirow{3}{*}{Yes} \\
    & FairMOT$^*$~\cite{zhang2021fairmot} & 73.7 & 72.3 & 43.2\% & 17.3\% & 3303 & \\
    & CorrTracker$^\ast$~\cite{wang2021multiple} & \textbf{76.5} & \textbf{73.6} & \textbf{47.6\%} & \textbf{12.7\%} & 3369 & \\
    \cline{2-8}
    %%%%%%%%%%%%%%%%%%%%%%%%%%%%%%%%%%%%%%%%%%%%%%%%%%%%%%%%%%%%%%%%%%%%%%%%%%%%%%%%%%%%%%%%%%%%%%%%
    & SST~\cite{sun2019deep} & 52.4 & 49.5 & 21.4\% & 30.7\% & 8431 & \multirow{3}{*}{No} \\
    & CenterTrack-MOTSynth$^*$~\cite{fabbri21motsynth} & 59.7 & 52.0 & - & - & 6035 &  \\
    & TDT-tracker$^*$ (Ours) & \textbf{63.8} & \textbf{60.9} & \textbf{35.4\%} & \textbf{17.3\%} & \textbf{4401} & \\
    % & TDT-tracker$^*$ (Ours) & \textbf{62.3} & \textbf{63.8} & 28.66\% & 25.61\% & \textbf{2607} & \\
    \midrule
    %%%%%%%%%%%%%%%%%%%%%%%%%%%%%%%%%%%%%%%%%%%%%%%%%%%%%%%%%%%%%%%%%%%%%%%%%%%%%%%%%%%%%%%%%%%%%%%%
    % \multirow{4}{*}{MOT20} & FairMOT~\cite{zhang2021fairmot} & 61.8 & 67.3 & \textbf{68.8\%} & \textbf{7.6\%} & 5243 & \multirow{2}{*}{Yes} \\
    \multirow{4}{*}{MOT20} & GSDT~\cite{Wang2021joint} & \textbf{67.1} & 67.5 & 53.1\% & 13.2\% & \textbf{3133} & \multirow{2}{*}{Yes} \\
    & CorrTracker$^*$~\cite{wang2021multiple} & 65.2 & \textbf{69.1} & \textbf{66.4\%} & \textbf{8.9\%} & 5183 & \\
    \cline{2-8}
    %%%%%%%%%%%%%%%%%%%%%%%%%%%%%%%%%%%%%%%%%%%%%%%%%%%%%%%%%%%%%%%%%%%%%%%%%%%%%%%%%%%%%%%%%%%%%%%%
    & Tracktor-MOTSynth~\cite{fabbri21motsynth} & 43.7 & 39.7 & - & - & \textbf{3467} & \multirow{2}{*}{No} \\
    & TDT-tracker$^*$ (Ours) & \textbf{47.9} & \textbf{46.0} & 18.5\% & 20.1\% & 5342 & \\
    \bottomrule
  \end{tabular}
  \caption{Comparison with existing trackers under the private MOT benchmark datasets~\cite{MOT16,MOTChallenge20}. These trackers are further categorized into the ones that use (\ie Yes in the last column) the expensive fully annotated real tracking data and the ones that do not (\ie No). The one-stage trackers are marked by `$^*$'.}
  \label{tab:compete_stoa}
\end{table*}

\subsubsection{MOT challenges}
We evaluate our TDT-Tracker on the private benchmark tracking datasets~\cite{MOT16,MOTChallenge20} to compare with some existing trackers (\cref{tab:compete_stoa}). There are two categories of trackers, namely the ones that use the fully annotated real tracking data in the training and the ones that do not. The state-of-the-art results are obtained by the former ones~\cite{zhang2021fairmot,Wang2021joint}.

From \cref{tab:compete_stoa}, our TDT-tracker outperforms other methods under the same category by a large margin on MOT17~\cite{MOT16} and MOT20~\cite{MOTChallenge20}. For example, TDT-tracker outperforms the second MOTA performance by 6.9\% (59.7 to 63.8) and the second IDF1 by 17.1\% (52.0 to 60.9) on MOT17. On MOT16 dataset, there are two two-stage trackers, namely CNNMTT~\cite{Mahmoudi2018MultitargetTU} and POI~\cite{yu2016poi}, that achieve better MOTA and IDF1 than TDT-tracker (we still outperform on Mostly Tracked Targets (MT) and Mostly Lost Targets (ML) metrics). CNNMTT uses the publicly available detections by a state-of-the-art detector and use the ID labels of MOT16 training set to train its embedder. Those ID labels are very useful since the MOT16 training and testing sets contain similar types of people appearing. POI~\cite{yu2016poi} uses different detection thresholds on different test sequences, while we only use one threshold. Adaptive thresholding helps achieve better performance on the benchmark datasets but it is not very practical. 

In addition, while maintaining very similar MOTA as JDE~\cite{wang2020JDE} on MOT16 benchmark, our TDT-tracker achieves 10.2\% better IDF1. JDE is a one-stage tracker and trained on some fully annotated tracking datasets.

\subsection{Ablation studies}
% Write an overview
% \raisebox{5.3mm}{Frame 2} & \includegraphics[width=0.224\textwidth]{images/occlusion/img2.png}

\subsubsection{Importance of teacher embedders}
% A plot shows difference of examples in the re-id datasets
% We expect that the joint model achieves better tracking performance with better teacher embedders. 
\setlength{\belowcaptionskip}{-4pt}
\begin{table}[h]
  \centering
  \begin{tabular}{@{}c|cccc@{}}
    \toprule
    Teacher Embedder & MOTA $\uparrow$ & IDF1 $\uparrow$ & IDs $\downarrow$ & AP $\uparrow$ \\
    \midrule
    ResNet50 (22.8) & 0.754 & 0.764 & 76 & 0.902 \\
    ON-Small (31.0) & 0.763 & 0.766 & 76 & 0.911 \\
    ON-Large (\textbf{43.3}) & \textbf{0.785} & \textbf{0.787} & \textbf{57} & \textbf{0.913} \\
    \bottomrule
  \end{tabular}
  \caption{Detection and tracking performance of our TDT-tracker with different teacher embedders. All these models use the same architecture and training settings. The number in the parentheses adjacent to each teacher embedder is its mean Average Precision (mAP) evaluated on the Market1501 dataset~\cite{zheng2015scalable}.}
  \label{tab:teacher_embedders}
\end{table}

\begin{figure}[h]
    \centering
        \begin{tabular}{cc|cc|cc}
        \toprule
        \multicolumn{2}{c|}{MSMT17} & \multicolumn{2}{c|}{Sunnyday} & \multicolumn{2}{c}{KITTI-13} \\
        \midrule
        \includegraphics[width=0.055\textwidth]{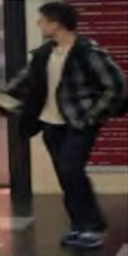} & \includegraphics[width=0.055\textwidth]{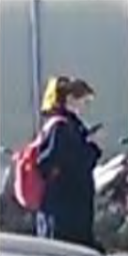} & \includegraphics[width=0.055\textwidth]{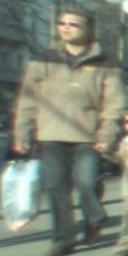} & \includegraphics[width=0.055\textwidth]{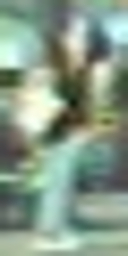} & \includegraphics[width=0.055\textwidth]{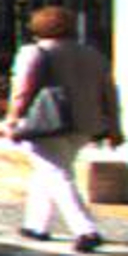} & \includegraphics[width=0.055\textwidth]{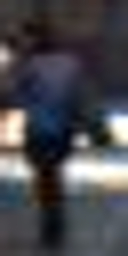} \\
        (a) & (b) & (c) & (d) & (e) & (f) \\
        \bottomrule
        \end{tabular}
    \caption{Left: Two examples from MSMT17 Re-ID dataset~\cite{wei2018person} that our teacher embedder is trained on. Middle and right: Examples from the Sunnyday sequence and KITTI-13 sequence of the MOT15~\cite{MOTChallenge2015} training set that is used as the validation set in ablation studies. (a) is a typical sample in the MSMT17 dataset, and (b) is an occluded sample. (c) and (e) are two clear examples from Sunnyday and KITTI-13 respetively, and (d) and (f) are unclear ones. The clear examples are closer to the ones in the MSMT17, and they are more common in the Sunnyday sequence but not in the KITTI-13 sequence.}
    \label{fig:train_sample_comparison}
\end{figure}

The teacher embedder determines the embedding quality of our joint model as it provides the pseudo labels for training. In \cref{tab:teacher_embedders}, we compare our TDT-tracker under three different teacher embedders, namely ResNet-50~\cite{he2016deep,torchreid}, OSNet-Small~\cite{zhou2019osnet,zhou2021osnet}, OSNet-Large~\cite{zhou2019osnet,zhou2021osnet}. These three embedders are trained on the MSMT17 Re-ID dataset~\cite{wei2018person} (\cref{fig:train_sample_comparison} left block) and achieve increasing mean Average Precision (mAP) on the Market1501 Re-ID dataset~\cite{zheng2015scalable}. 

% Higher mAP indicates the teacher embedder can generate more discriminative embeddings. 

Interestingly, the one-stage tracker achieves increasing detection performance with better teacher embedders, and the largest AP increase is $1.2\%$ (0.902 to 0.913). The tracking performance also increases with better teacher embedders. For example, comparing to using ResNet50 as the teacher embedder, TDT-tracker achieves 4.1\% better MOTA and 25.0\% lower ID switches. 

% Teacher embedder performance on Market1501 dataset: https://kaiyangzhou.github.io/deep-person-reid/MODEL_ZOO
\setlength{\belowcaptionskip}{-8pt}

\setlength{\belowcaptionskip}{-4pt}
\begin{table}[h]
  \centering
  \begin{tabular}{@{}l|cc|cc@{}}
    \toprule
    \multirow{2}{*}{Dataset} & \multicolumn{2}{c|}{FairMOT}  & \multicolumn{2}{c}{TDT-RNet34} \\
    & AP $\uparrow$ & MOTA $\uparrow$ & AP $\uparrow$ & MOTA $\uparrow$ \\
    \midrule
    Overall & \textbf{0.932} & \textbf{80.5} & 0.926 & 78.4 \\
    KITTI13 & \textbf{0.796} & \textbf{42.7} & 0.792 & 32.3 \\
    ETH-SunnyDay & \textbf{0.997} & 84.3 & 0.984 & \textbf{88.7} \\
    \bottomrule
  \end{tabular}
  \caption{Comparison between FairMOT~\cite{zhang2021fairmot} and TDT-tracker on the detection and tracking performance separately. Both models are trained on the same datasets including the BaseTrainSet and CrowdHuman. KITTI13 and ETH-Sunnyday are two sequences in the validation set.}
  \label{tab:compare_det_emb}
\end{table}

% Particulary, our TDT model achieves comparable or better results on videos with large and clear people, but fails on small and largely overlapped cases.

% Most of the pedestrian image in the ETH-Sunnyday (middle block in \cref{fig:train_sample_comparison}) looks closer to the training samples in the MSMT17 dataset (left block) where each identity is clear.
% From first row in \cref{fig:train_sample_comparison}, FairMOT outperforms TDT-tracker on the overall tracking performance (MOTA). Digging into more details.

However, the current teacher embedder cannot support our joint model to achieve comparable tracking performance as the state-of-the-art tracker~\cite{zhang2021fairmot,wang2021multiple,Wang2021joint} (trained on the fully annotated tracking labels) on some cases. In \cref{tab:compare_det_emb}, we compare the performance of a FairMOT~\cite{zhang2021fairmot} model and our TDT model on the validation dataset. Both models have been trained on the same datasets, namely the BaseTrainSet and the CrowdHuman dataset, and use the ResNet-34 as the backbone network. In a sequence named KITTI13 (second row in \cref{tab:compare_det_emb}) from the validation set, the most of the pedestrians appearing in the video are not clear (see \cref{fig:train_sample_comparison} (f)), and our embedder does much worse (24.36\% worse on MOTA) than FairMOT. In contrast, our embedder outperform FairMOT by 5.22\% on MOTA on the sequence ETH-Sunnyday (third row) where most people are large and clear in the video (see \cref{fig:train_sample_comparison} (c)). Both detectors perform comparably on both sequences, and thus, the embedding quality is the main cause. FairMOT does better on the KITTI13 likely attributes to their privilege of seeing similar samples during training. However, this privilege may not always exist.

\begin{figure*}[h]
    \centering
        \begin{tabular}{c|cccc}
        & Frame T & Frame T + 100 & Frame T + 200 & Frame T + 300 \\
        \midrule
        \raisebox{7mm}{(a)} & \includegraphics[width=0.20\textwidth]{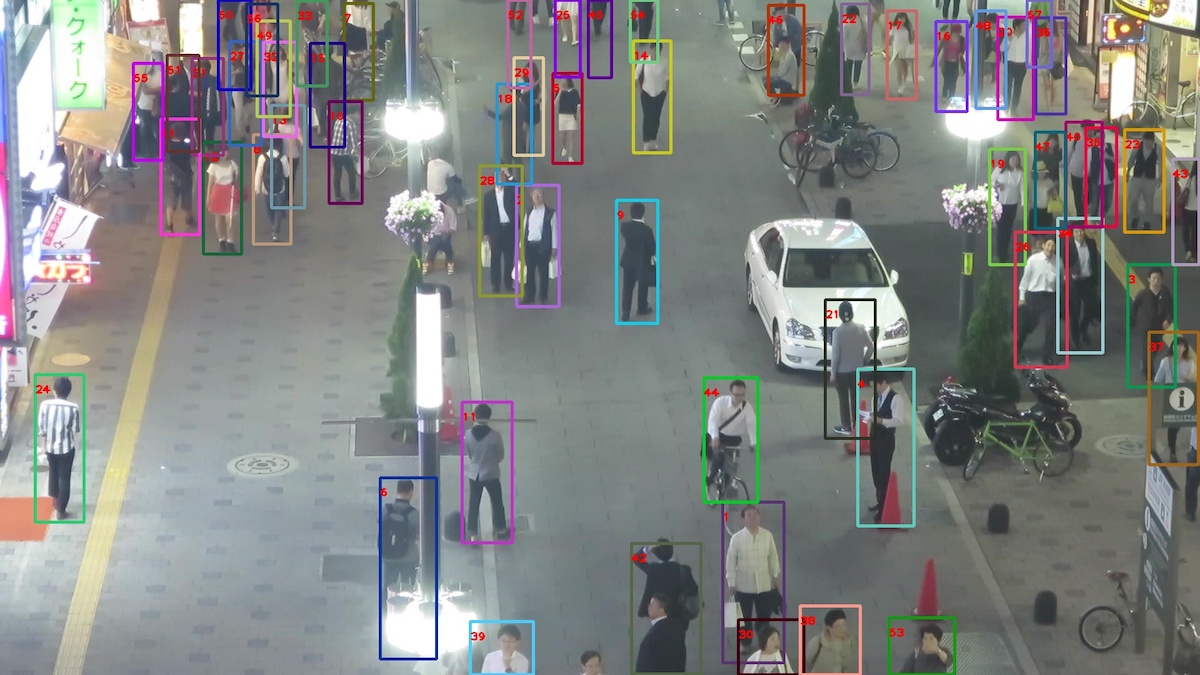} & \includegraphics[width=0.20\textwidth]{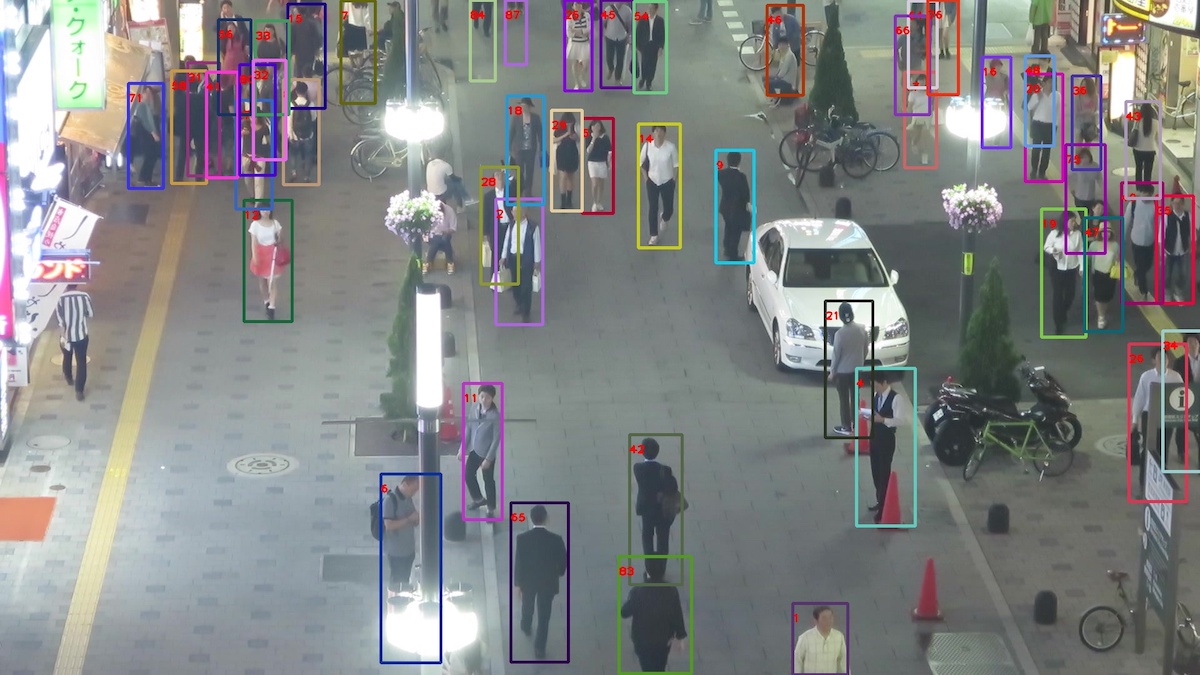} & \includegraphics[width=0.20\textwidth]{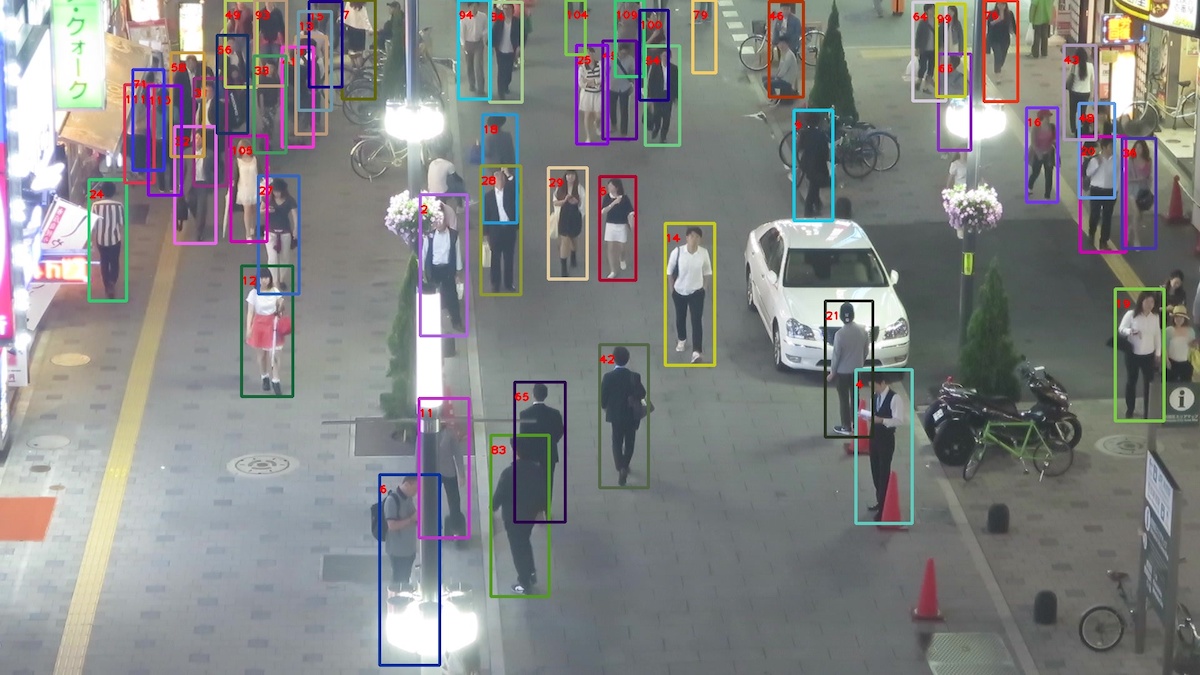} & \includegraphics[width=0.20\textwidth]{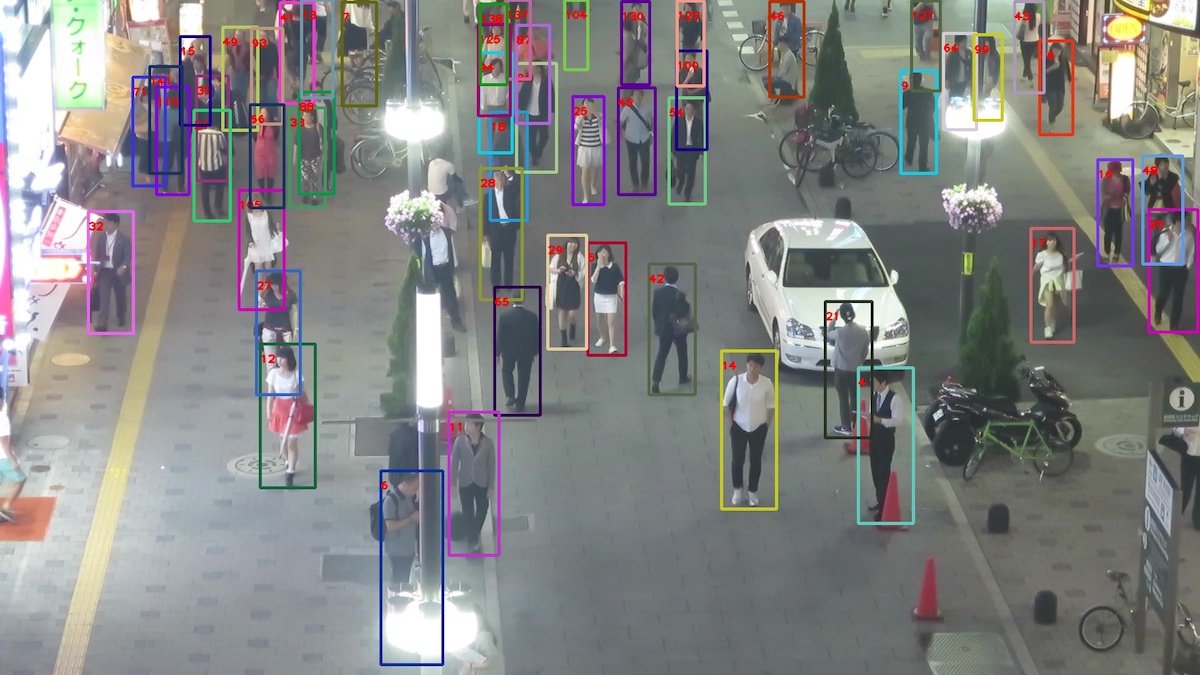} \\
        \raisebox{7mm}{(b)} & \includegraphics[width=0.20\textwidth]{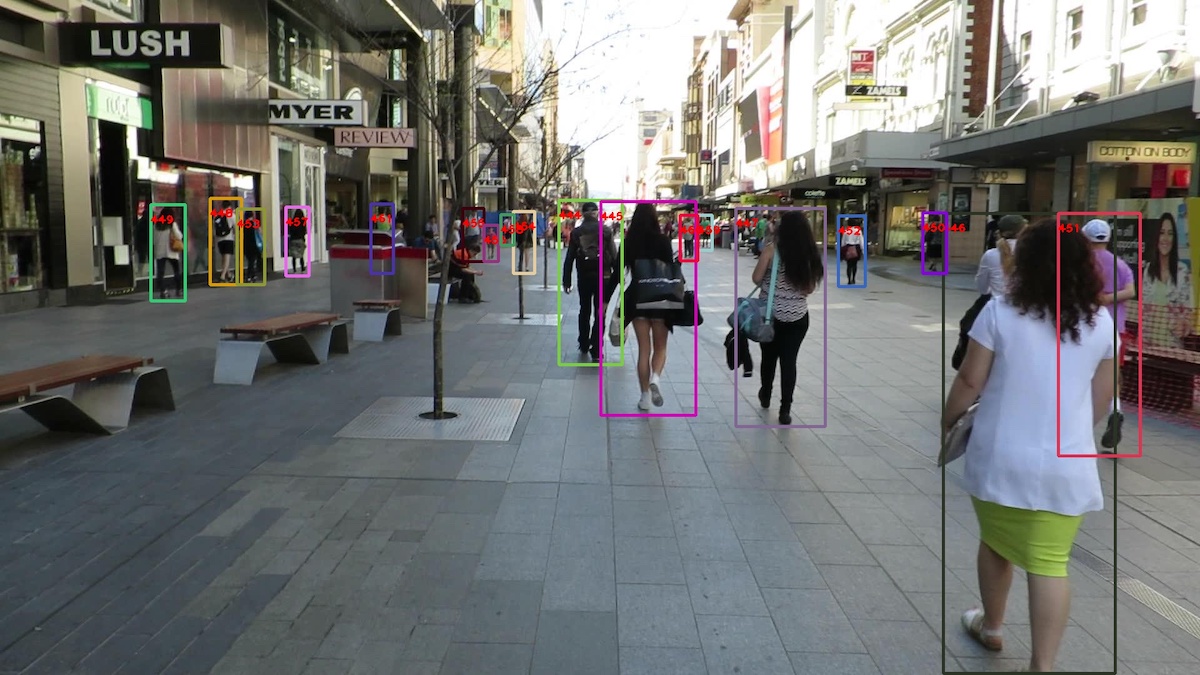} & \includegraphics[width=0.20\textwidth]{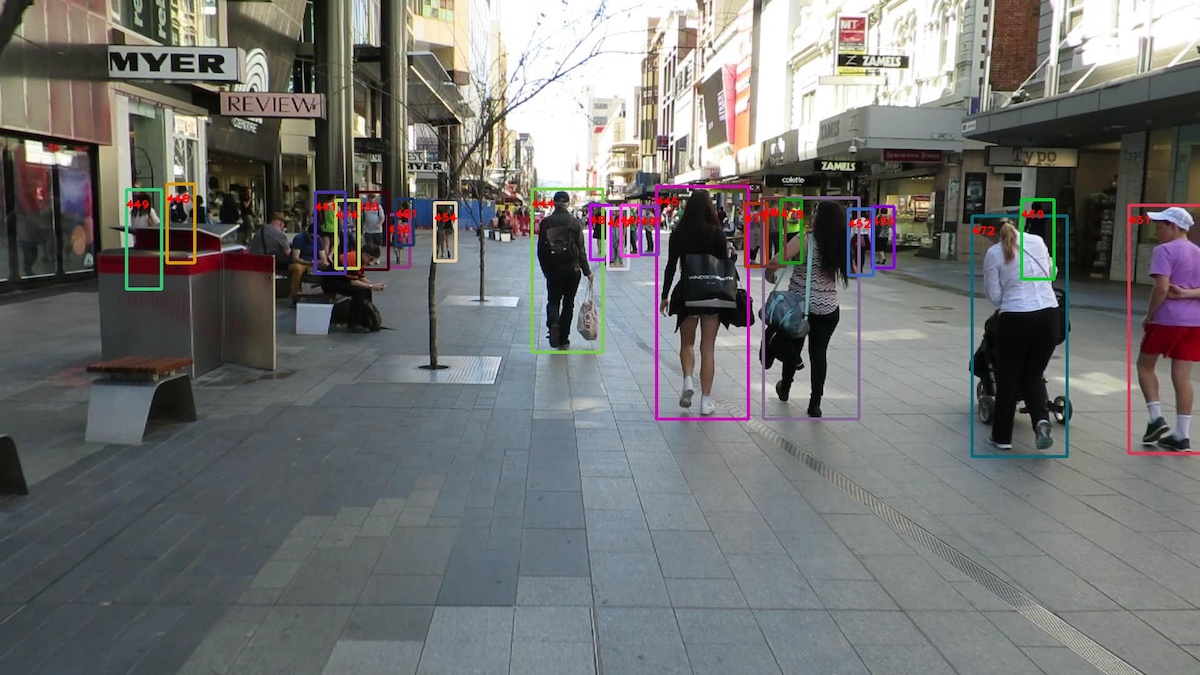} & \includegraphics[width=0.20\textwidth]{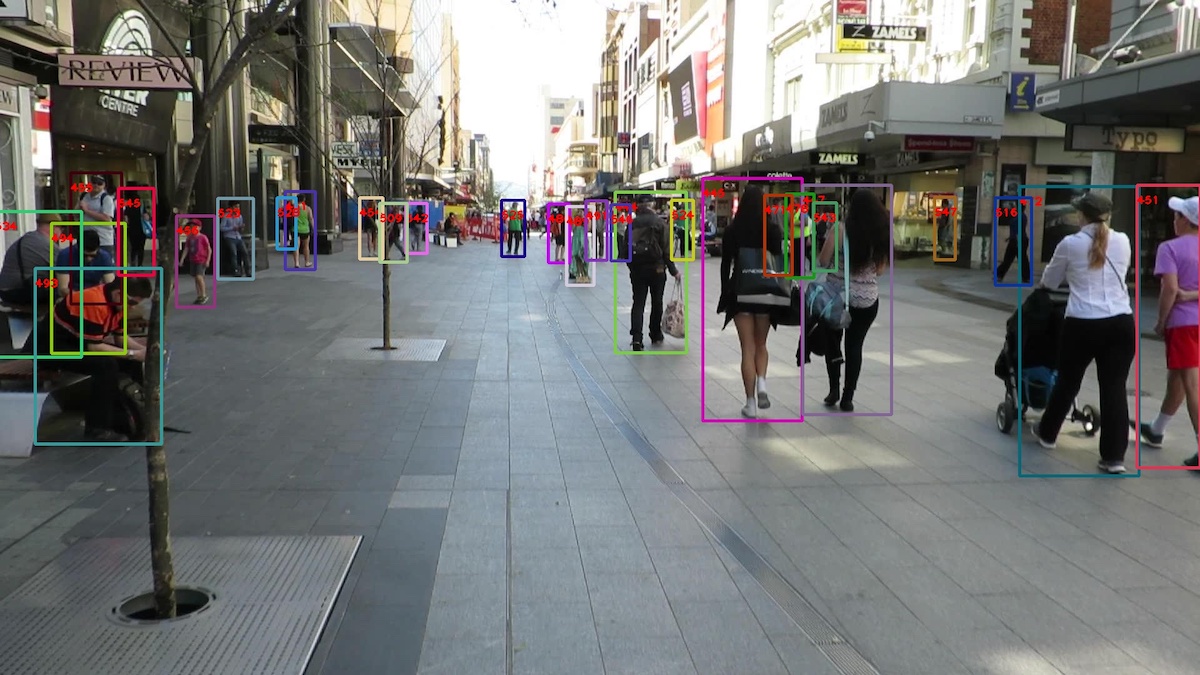} & \includegraphics[width=0.20\textwidth]{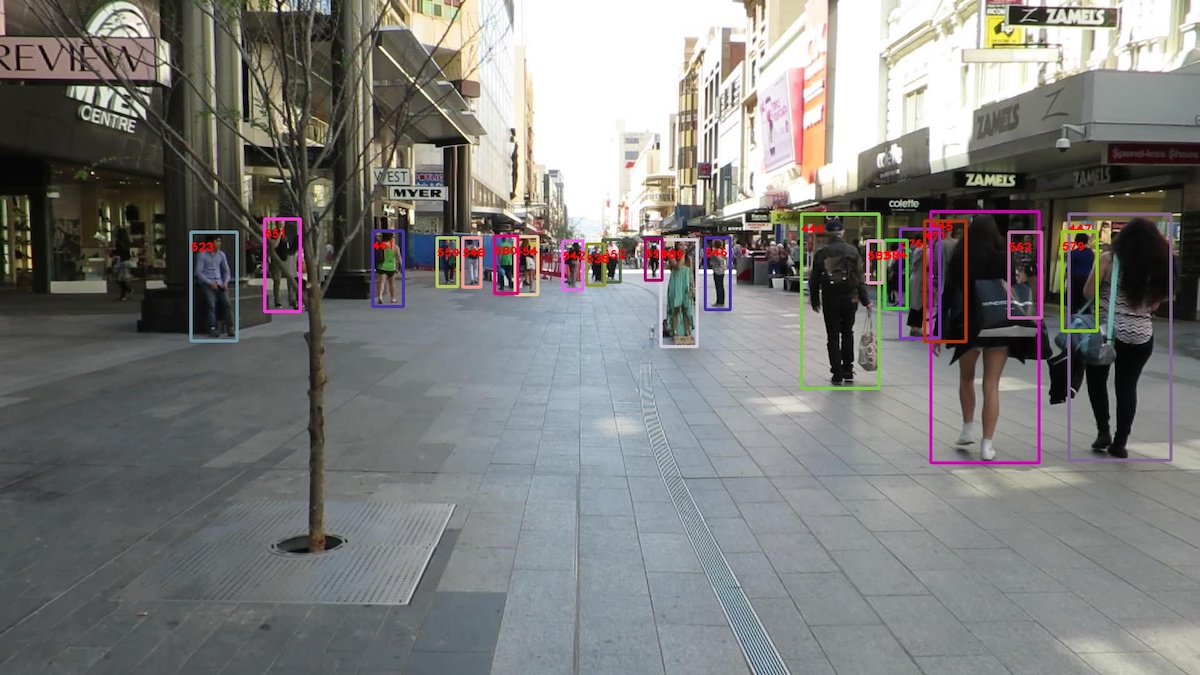} \\
        \raisebox{7mm}{(c)} & \includegraphics[width=0.20\textwidth]{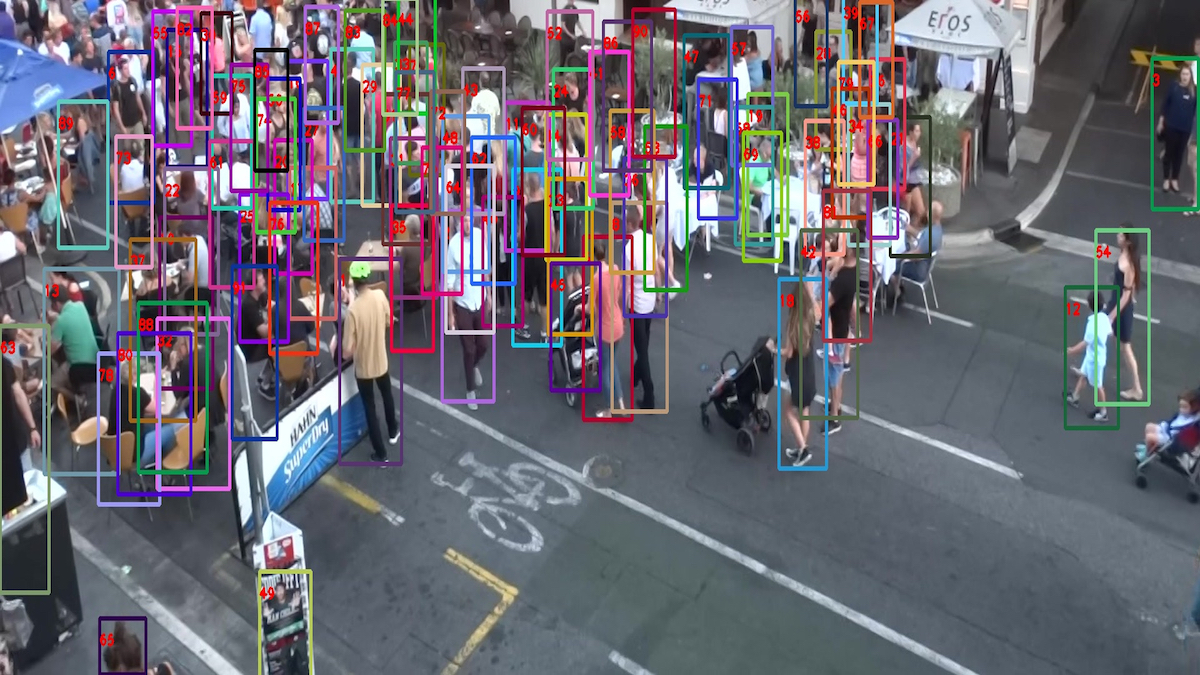} & \includegraphics[width=0.20\textwidth]{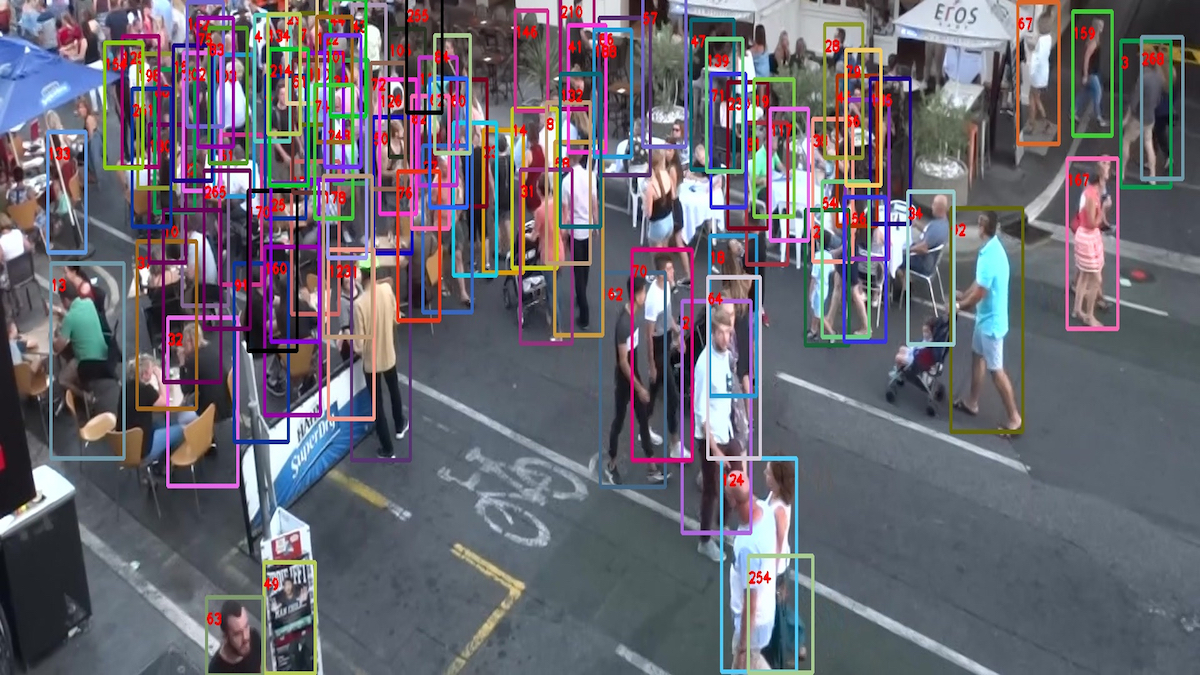} & \includegraphics[width=0.20\textwidth]{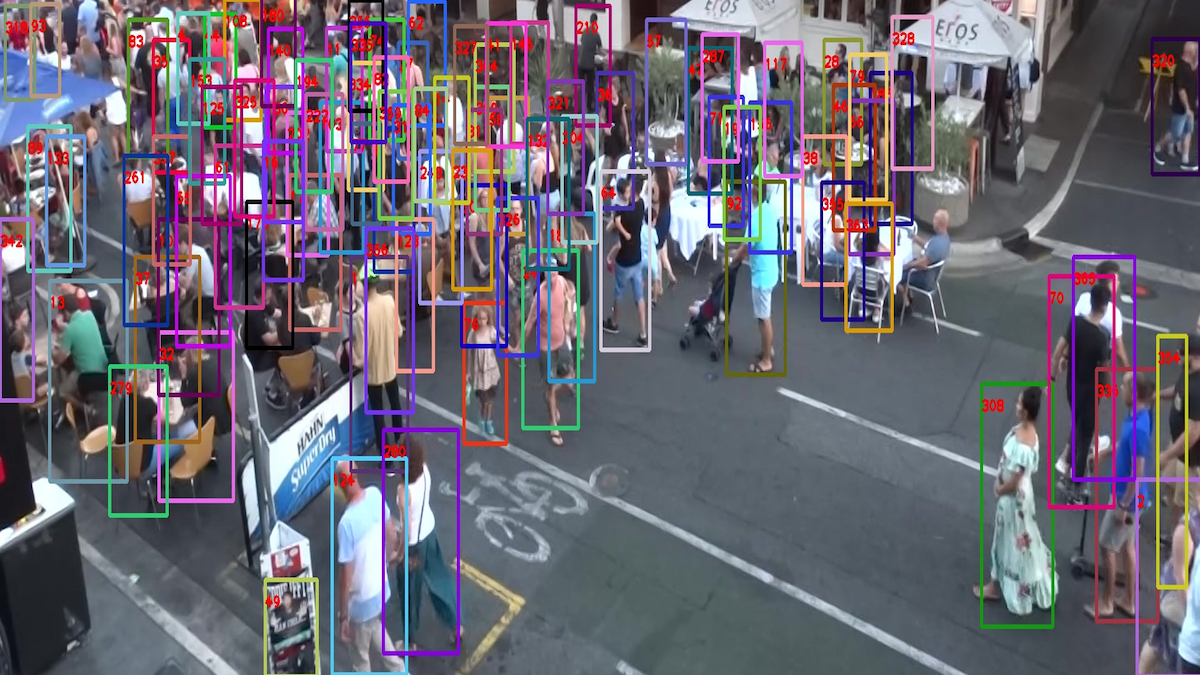} & \includegraphics[width=0.20\textwidth]{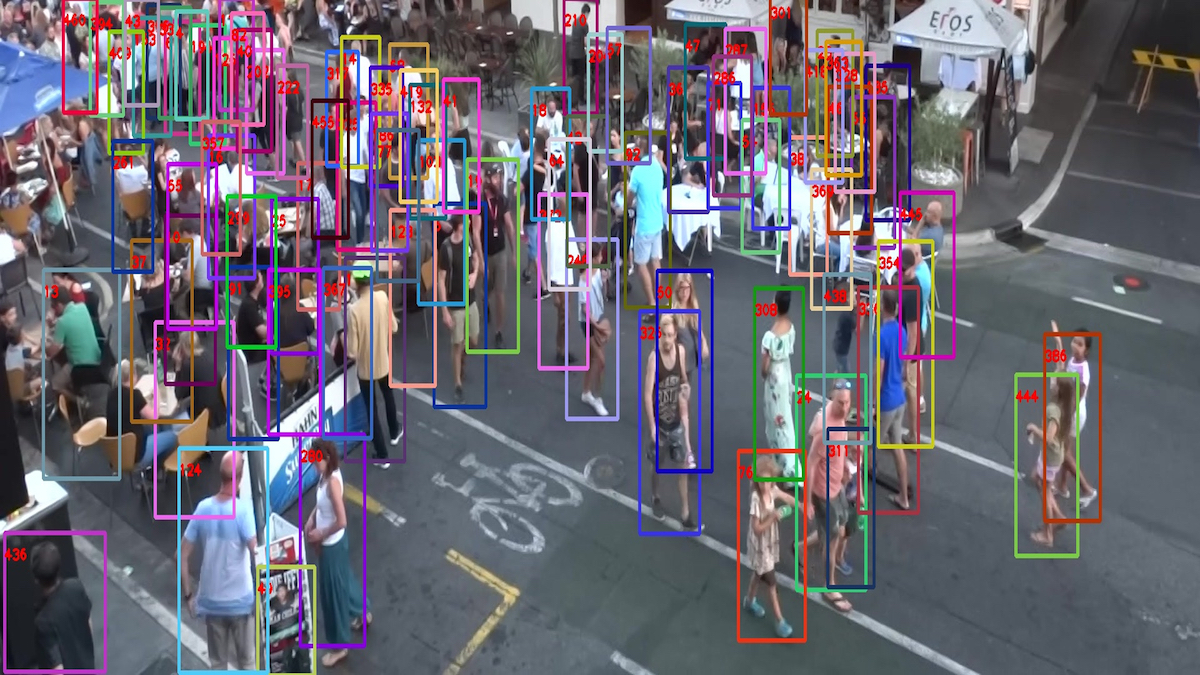} \\
        % \raisebox{7mm}{(d)} & \includegraphics[width=0.20\textwidth]{images/MOT17-14/00400.jpg} & \includegraphics[width=0.20\textwidth]{images/MOT17-14/00500.jpg} & \includegraphics[width=0.20\textwidth]{images/MOT17-14/00600.jpg} & \includegraphics[width=0.20\textwidth]{images/MOT17-14/00700.jpg} \\
        \end{tabular}
    \caption{Three qualitative examples by our TDT-tracker on video sequences~\cite{MOT16,MOTChallenge20} of different types. Sequence (a) and (b) contain high-resolution people with less occlusions. Sequence (c) shows the crowded scenario.}
    \label{fig:qualitative_analysis}
\end{figure*}

\Cref{fig:qualitative_analysis} shows qualitative results from our TDT-tracker on different types of video sequences in MOT17~\cite{MOT16} and MOT20~\cite{MOTChallenge20} test sets. Most of the objects are accurately detected and our TDT-tracker can track well even on occluded people (see the successful tracking of the fifth person from left with ID 461 in the starting frame of sequence (a) in \cref{fig:qualitative_analysis}).

\subsubsection{Backbone choices for the joint model}
\setlength{\belowcaptionskip}{-6pt}
\begin{table}[h]
  \centering
  \begin{tabular}{@{}l|cccccc@{}}
    \toprule
    Model & MOTA$\uparrow$ & IDF1$\uparrow$ & IDs$\downarrow$ & AP$\uparrow$ & FPS$\uparrow$ \\
    \midrule
    MNet & 0.744 & 0.747 & 72 & 0.907 & \textbf{11.18} \\
    MNet-FPN & 0.764 & 0.759 & 73 & 0.910 & 10.01 \\
    RNet34-FPN & \textbf{0.785} & \textbf{0.787} & \textbf{57} & \textbf{0.913} & 9.61 \\
    RNet50-FPN & 0.757 & 0.786 & 58 & 0.912 & 7.93 \\
    \bottomrule
  \end{tabular}
  \caption{Impacts of different backbone architecture on the performance of our TDT-tracker. MNet refers to MobileNet~\cite{Howard17mobilenets} and RNet is ResNet~\cite{he2016deep}.}
  \label{tab:arch_choices}
\end{table}

A suitable backbone architecture is essential to the success of our weakly supervised scheme. We need the backbone features to contain not only the information for detecting objects but also information on the distinguishing properties of the objects. In addition, these properties can be extracted for people of different sizes in an image. In \cref{tab:arch_choices}, we show the performance of detection and tracking under different choices of backbone architecture. 

% We use OSNet-Large as the teacher embedder for these experiments.

From \cref{tab:arch_choices}, we can see that the feature fusion mechanism helps the overall tracking performance. With the FPN, MOTA improves 2.7\% (0.744 to 0.764) on the MobileNet backbone, \ie from MNet to MNet-FPN. A more powerful backbone architecture boosts the tracking performance by another 2.75\% (0.764 to 0.785), \ie from MNet-FPN to RNet34-FPN. The improved embedding quality is the main reason since the detection results are almost the same but the number of ID switches drops $21.9\%$ (from 73 to 57). Interestingly, using ResNet-50 decreases the overall tracking performance measured by MOTA by a large margin. The reason is that the tracking precision (as in the CLEAR metrics~\cite{bernardin2008evaluating}) decreases while other metrics stay similar. Similar phenomenon is also observed in FairMOT~\cite{zhang2021fairmot}, where they also find that simply increasing the backbone power may not improve the overall tracking performance.

% Sequence (a) contains clear people with limited overlapping, which are close to the samples in the dataset MSMT17~\cite{wei2018person} that our teacher embedder is trained on. Our TDT-tracker is able to track most of them with very low ID switches~\cref{tab:compete_stoa}. Particularly, our TDT-tracker successfully tracks the person in green shirt (the fifth person from left on the starting frame with ID label 461) across all frames. This is an encouraging example of our proposed framework because that the lower body of that person is blocked at the starting frame, and yet the embedding can caputure the distinguishing features out of it. Similar observations also hold for people with ID 448 and 449, where their lower body is blocked after 100 frames. This likely attributes to the occluded examples (\cref{fig:train_sample_comparison} (b)) in the MSMT17 Re-ID dataset that our teacher embedder is trained on. In comparison, people in sequence (b) commonly overlap with each other and TDT-tracker tracks them poorly.

\subsubsection{Impact of embedding dimensionality}
\begin{table}[h]
  \centering
  \begin{tabular}{@{}l|cccccc@{}}
    \toprule
    Dim & MOTA $\uparrow$ & IDF1 $\uparrow$ & IDs $\downarrow$ & AP $\uparrow$ & FPS $\uparrow$ \\
    \midrule
    64 & 0.739 & 0.772 & 67 & 0.902 & \textbf{13.34} \\
    128 & 0.773 & 0.782 & \textbf{55} & 0.908 & 11.82 \\
    256 & 0.762 & 0.777 & 63 & 0.910 & 10.76 \\
    512 & \textbf{0.785} & \textbf{0.787} & 57 & \textbf{0.913} & 9.61 \\
    \bottomrule
  \end{tabular}
  \caption{The effects of using different portion of the teacher embedding as the supervision signal. All the models share the same architecture except for the last layer of the embedding head since embeddings of different dimensions are generated.}
  \label{tab:dim_analysis}
\end{table}

A practical issue of using the proposed weakly supervised framework is that the dimension of the teacher embedder can be too large, which can create a large memory and computational overhead, especially for the anchor based detection models as they predict an embedding for each anchor box.

As mentioned in \cref{sec:losses}, a simple way to circumvent the situation is to use the first $D_s$ elements of the teacher embedding. Recall that $D_s$ represents the size of the desired output embedding, \ie student embedding. \cref{tab:dim_analysis} shows the performance of setting $D_s$ to be 64, 128, 256, and 512 respectively, given that $D_t$ (dimension of the teacher embedding) is 512. First, there is a clear pattern of improving detection performance as larger portion of the teacher embedding is used. Interestingly, in comparison to the full teacher embeddings, our joint model can achieve 94.1\% (0.739 out of 0.785) of the tracking performance measured in MOTA and $98.1\%$ if measured in IDF1 by using only one eighth of them. However, increasing the embedding portion generally leads to better tracking performance and using the full teacher embedding still bests other settings. More dimension contains more information that can be helpful in discriminating people. In addition, the teacher embedding may not be evenly distributed, and the first part likely cannot represent the full information. Note that the $D_s = 128$ and $D_s = 256$ are against the trend, where the tracking performance decreases as $D_s$ increases. We believe this is due to the randomness of data. On the other hand, there is a clear increasing pattern of the inference speed as $D_s$ decreases. Thus, this is a typical speed and accuracy trade off situation.

\subsubsection{Training data}
\setlength{\belowcaptionskip}{-8pt}
\begin{figure}[h]
    \centering
    \includegraphics[width=0.5\linewidth]{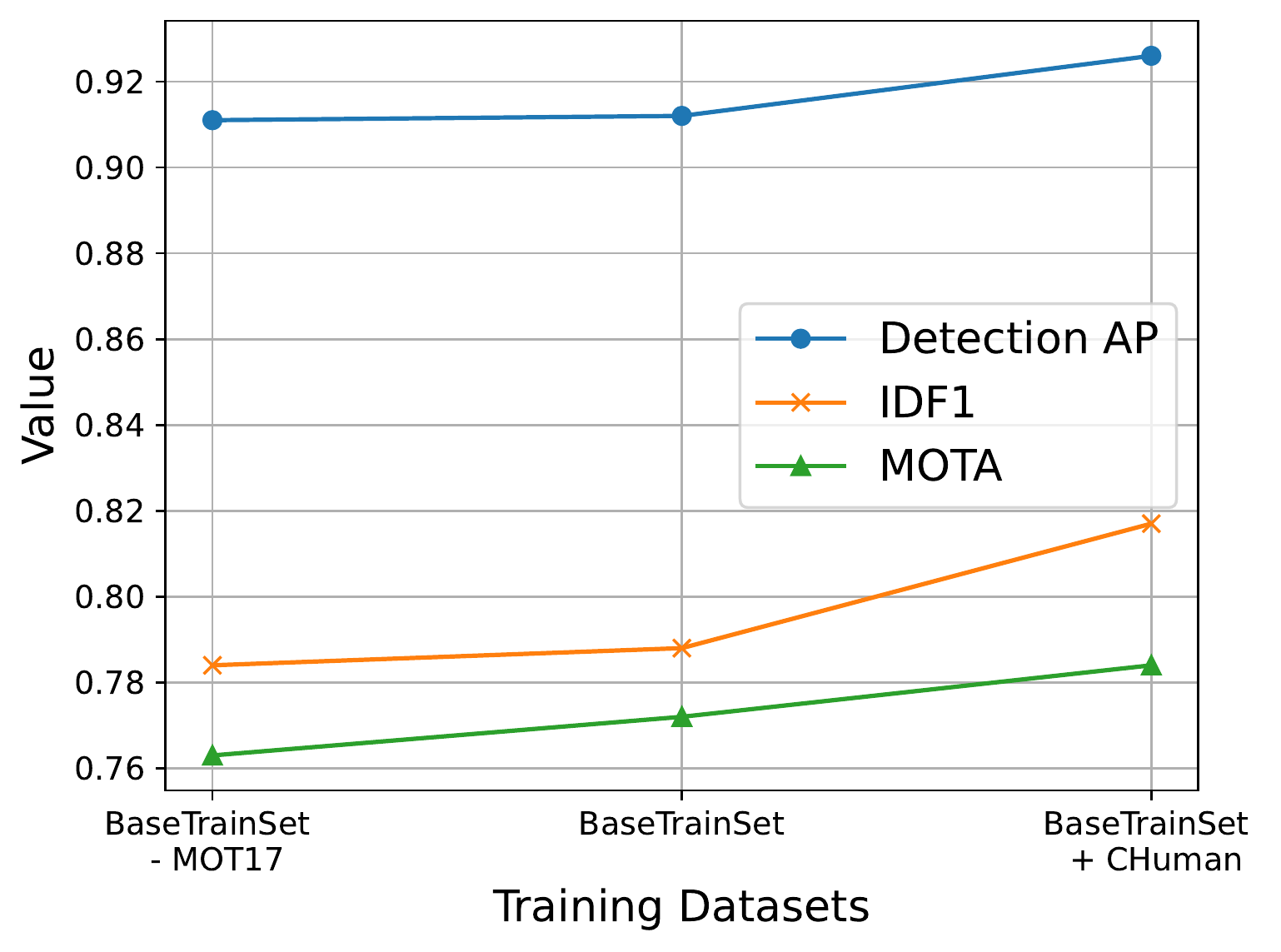}
    \caption{Improvement of the performance in both the detection (AP) and tracking (MOTA and IDF1) of our proposed system as more distilled detection datasets are used for training. These three models share the same architecture and training settings. The higher the better for these metrics.}
    \label{fig:dataset_impact}
\end{figure}

Under the proposed weakly supervised framework, any detection dataset can be easily augmented with some teacher embeddings and used as a tracking dataset to train a joint model for tracking. This is an advantage. We show in \cref{fig:dataset_impact} that the overall tracking performance, measured in both MOTA and IDF1, improves as more augmented datasets are used in training. The detector also improves, although marginally, as well. This means our weakly supervised framework benefits from the diversity of the detection datasets, which is promising because detection datasets are easier to annotate than tracking datasets. 

% \begin{table}
%   \centering
%   \begin{tabular}{@{}l|ccccc@{}}
%     \toprule
%     Datasets & MOTA$\uparrow$ & IDF1$\uparrow$ & IDs$\downarrow$ & AP$\uparrow$ \\
%     \midrule
%     BaseTrainSet - MOT17 & 0.763 & 0.784 & 57 & 0.911 \\
%     BaseTrainSet & 0.772 & 0.788 & 56 &  0.912\\
%     BaseTrainSet + CHuman & 0.784 & 0.817 & 56 & 0.926 \\
%     \bottomrule
%   \end{tabular}
%   \caption{}
%   \label{tab:dataset_impact}
% \end{table}

%------------------------------------------------------------------------
\section{Conclusion}
\label{sec:disc_con}
In this paper, we propose a simple but effective embedding distillation framework that aims to mitigate the issue of expensive and scarce fully annotated real tracking data. Our TDT-tracker achieves competitive results on the benchmark datasets and we analyze various aspects of our framework. Particularly, we have shown that our tracker achieves better performance with better embedders. This is a promising direction since better Re-ID networks are actively being proposed, which directly improves our one-stage tracker. While our work has avoided using any tracking data to train the teacher embedder, it is still possible and practical to sample some pairs of frames from videos, have them annotated, and use their annotations as Re-ID examples to further improve the student tracker performance.

%%%%%%%%% REFERENCES
{\small
\bibliographystyle{ieee_fullname}
\bibliography{egbib}
}

\newpage
\section*{Supplementary Materials}
\begin{appendix}

\section{Additional implementation details}
We trained our joint model on the BaseTrainSet and CrowdHuman datasets for 200K iterations with batch size 32 and evaluated it on the MOT16 and MOT17 private test sets. We further fine-tuned this model on the MOT20 training set for another 30K iterations with batch size 16 and evaluated it on the MOT20 test set. All of our models are trained from scratch without pre-training on any datasets.

We set $\epsilon$ as 0.001 and the momentum as 0.997 for our batch normalization layer. The $\gamma$ and $\beta$ are learnable.

The L2 weight decay was set 0.0001. We applied the cosine learning rate decay during training and the initial learning rate was 0.15. We warmed up the training process with a small learning rate 0.001 for the first 2K iterations. During fine-tuning, we set the initial learning rate as 0.015 with the same warming up procedure.

The detection threshold was 0.5 for MOT16 and MOT17 test sets and 0.3 for MOT20 test set.

Our system was implemented in TensorFlow. Our current implementation is not optimal and the running speed would be faster with more careful design and implementation.

\section{More ablation studies on parameters}
\begin{figure}[h]
    \centering
        \includegraphics[width=0.40\textwidth]{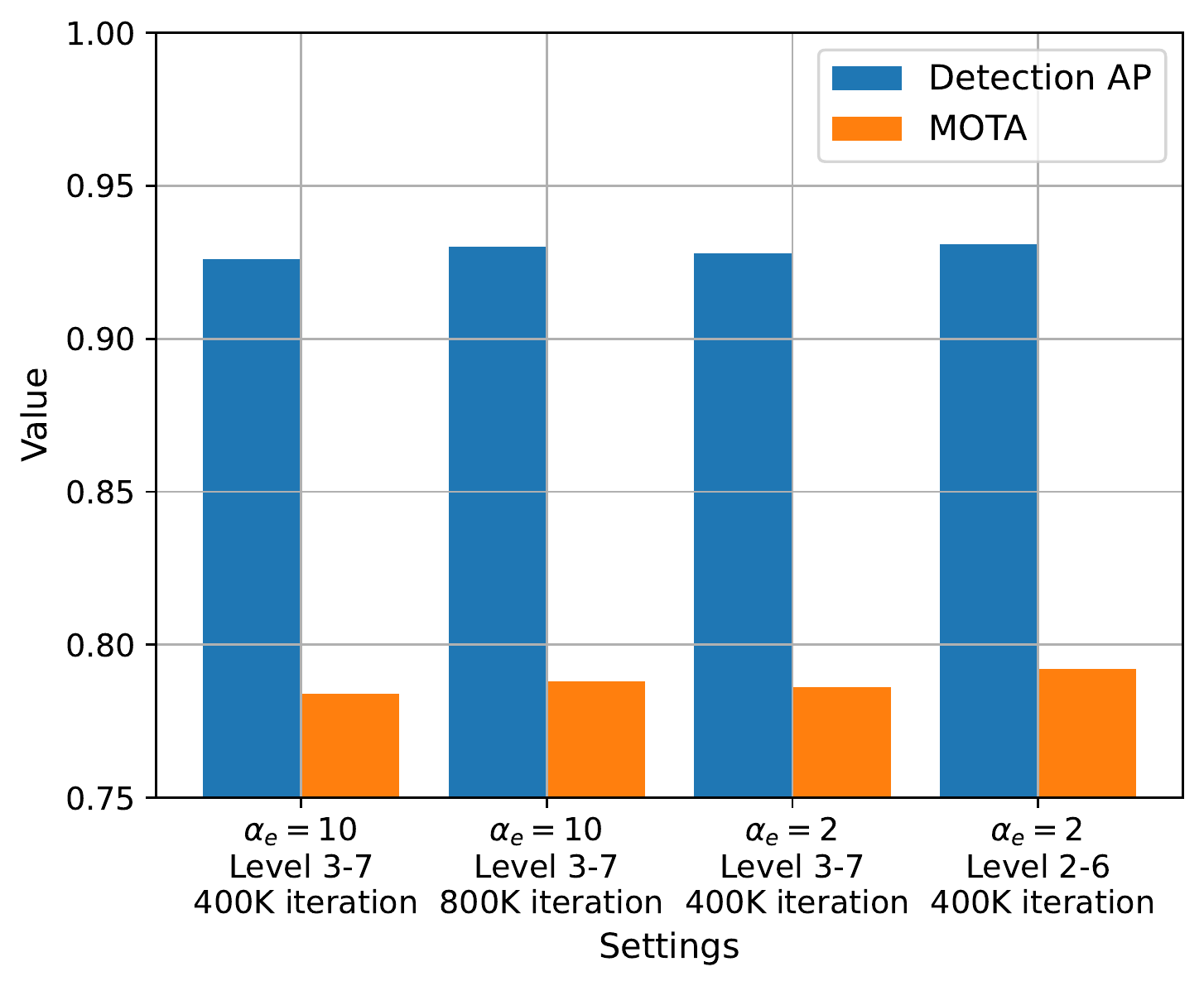}
    \caption{Comparison of detection (Average Precision) and tracking (MOTA) among different hyper-parameter settings. All these models have the same architecture. The backbone architecture is ResNet-34.}
    \label{fig:more_ablation}
\end{figure}

\Cref{fig:more_ablation} shows the ablation studies on some other hyper-parameters of our proposed system, namely the training time, pyramid levels of FPN, and the weight of embedding loss. From the figure, we see that doubling the training time only marginally helps the detection and tracking performance (comparing group 1 and 2). In addition, our system is robust to the weight of the embedding loss. Both the detection and tracking have similar performance between setting $\alpha_e = 2$ and $\alpha_e = 10$ (comparing group 1 and 3). Using features of larger resolutions from the Feature Pyramid Network does not have much impact either (comparing group 3 and 4).

\section{Inference speed of two-stage tracker}
One of the advantages of the one-shot tracker is its faster inference speed than the two-stage tracker. Since the joint model generates an embedding for every anchor box in one forward pass, the running time stays the same regardless of the number of objects in the frame. However, it takes two-stage trackers increasing time with more objects in the scene. For example, the inference speed of the counterpart two-stage tracker of our TDT-tracker is 3.03 FPS for scenes with less people (\eg MOT15) and 0.98 FPS for a scene with more people (\eg sequence MOT20-05). In comparison, our one-shot tracker used very similar running time around 10 FPS regardless of the types of the scene.

\section{Detailed performance on the benchmark datasets}
\begin{figure*}[h]
    \centering
        \includegraphics[width=0.90\textwidth]{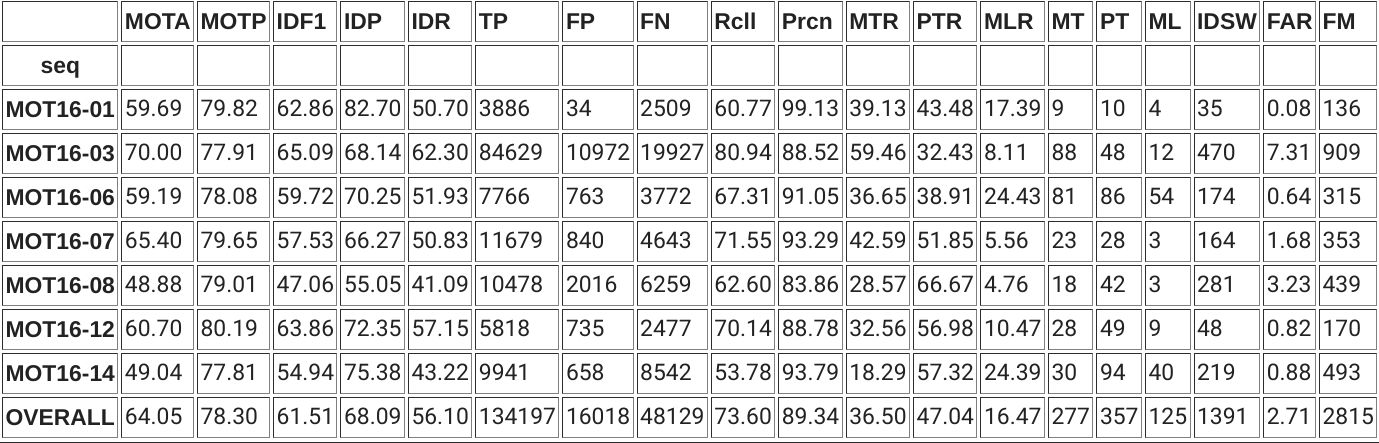}
    \caption{Detailed tracking performance of our TDT-tracker on each testing sequence in MOT16.}
    \label{fig:mot16}
\end{figure*}

\begin{figure*}[h]
    \centering
        \includegraphics[width=0.90\textwidth]{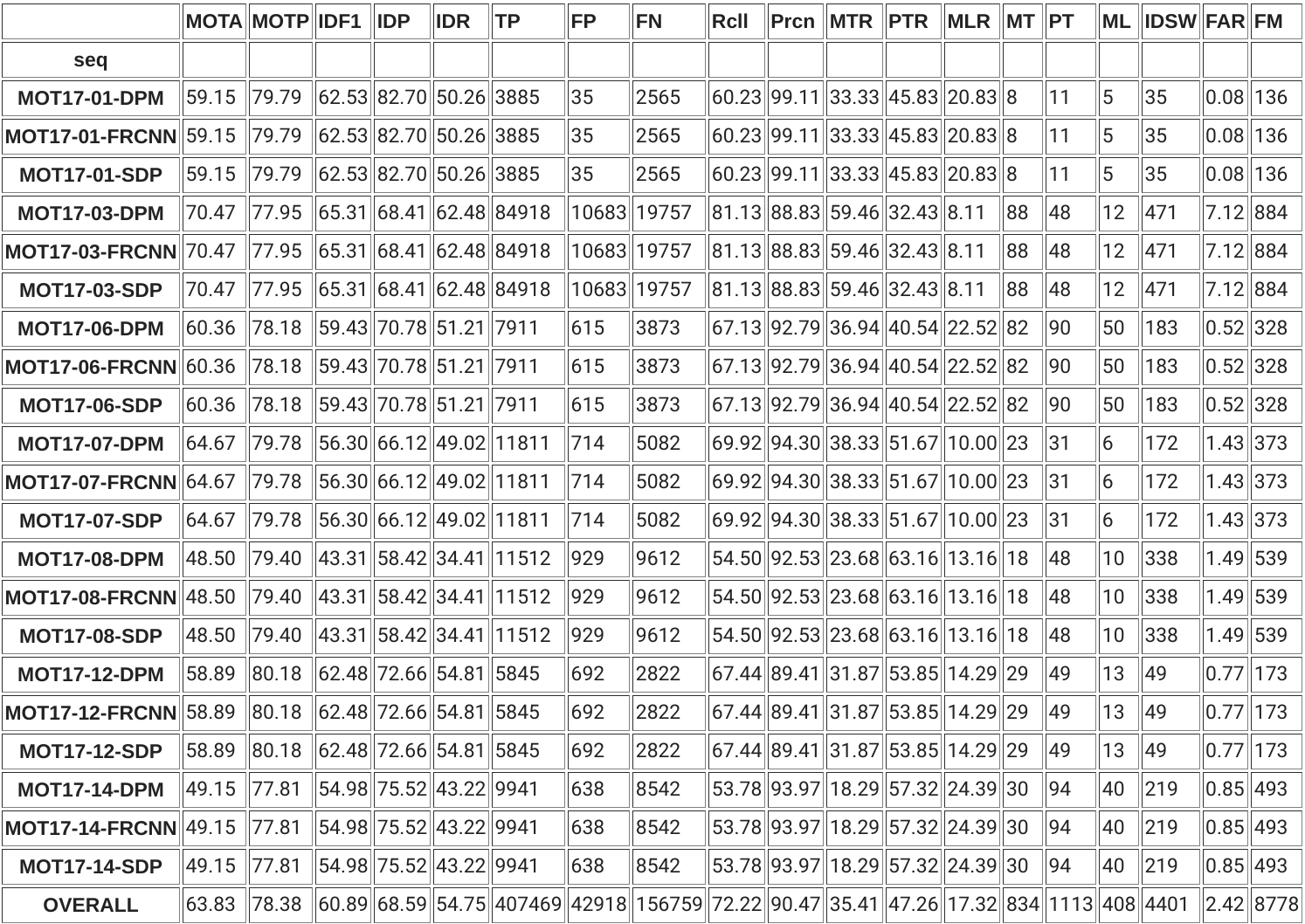}
    \caption{Detailed tracking performance of our TDT-tracker on each testing sequences in MOT17.}
    \label{fig:mot17}
\end{figure*}

\begin{figure*}[h]
    \centering
        \includegraphics[width=0.90\textwidth]{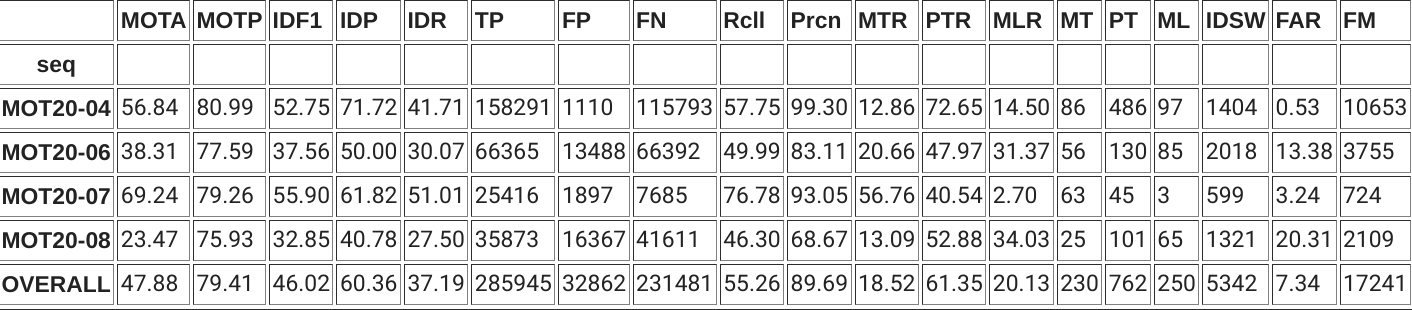}
    \caption{Detailed tracking performance of our TDT-tracker on each testing sequence in MOT20.}
    \label{fig:mot20}
\end{figure*}

\Cref{fig:mot16}, \cref{fig:mot17}, and \cref{fig:mot20} show the detailed tracking performance of our TDT-tracker on different video sequences of MOT16, MOT17, and MOT20 respectively. These resutls were evaluated by the private benchmark server. In general, our TDT-tracker does well on sequences with clear people but poorly on those crowded sequences or those with small people. State-of-the-art trackers have better performance due to their previlege of training on the fully annotated tracking datasets that are similar to these test datasets. Our TDT-tracker would largely improve if our teacher embedder sees similar samples during training.

\section{Qualitative analysis}
\begin{figure*}[h]
    \centering
        \begin{tabular}{cccc}
        Frame 1 & Frame 101 & Frame 201 & Frame 301 \\
        \midrule
        \includegraphics[width=0.22\textwidth]{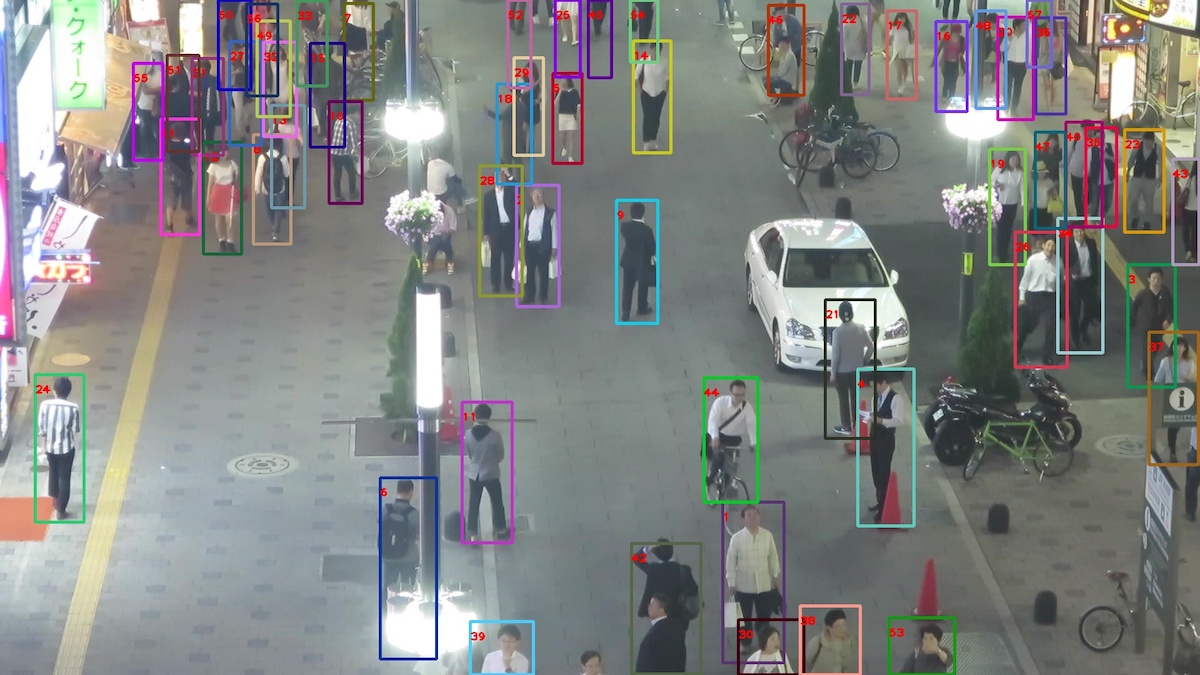} & \includegraphics[width=0.22\textwidth]{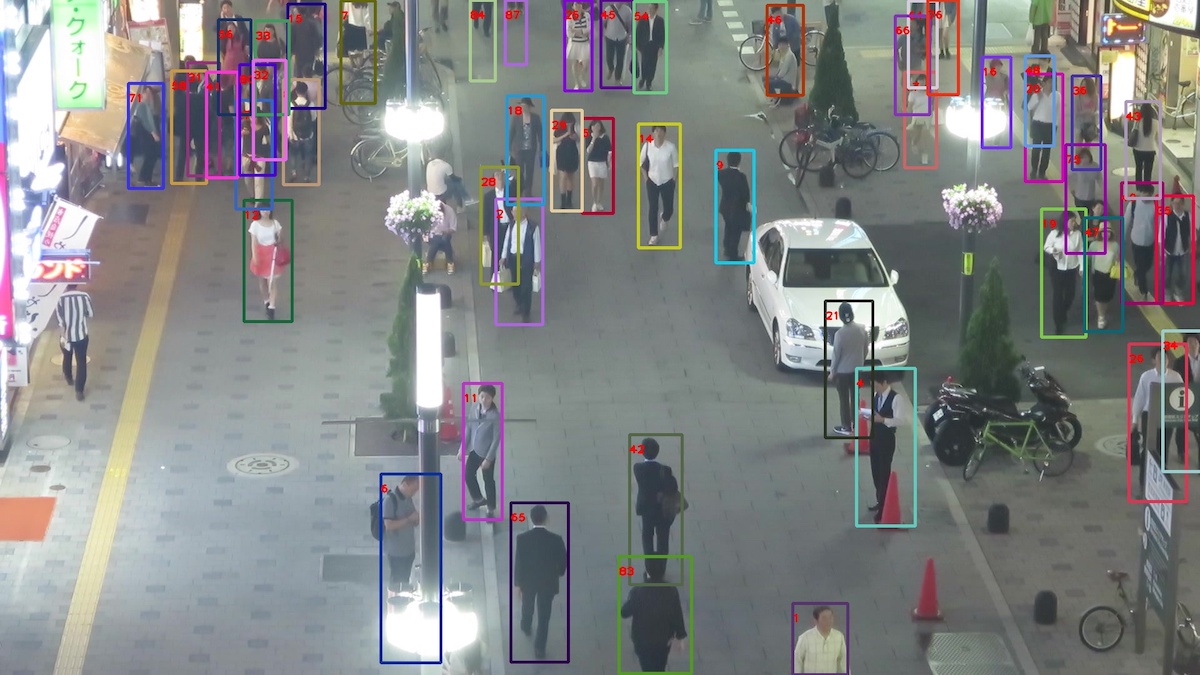} & \includegraphics[width=0.22\textwidth]{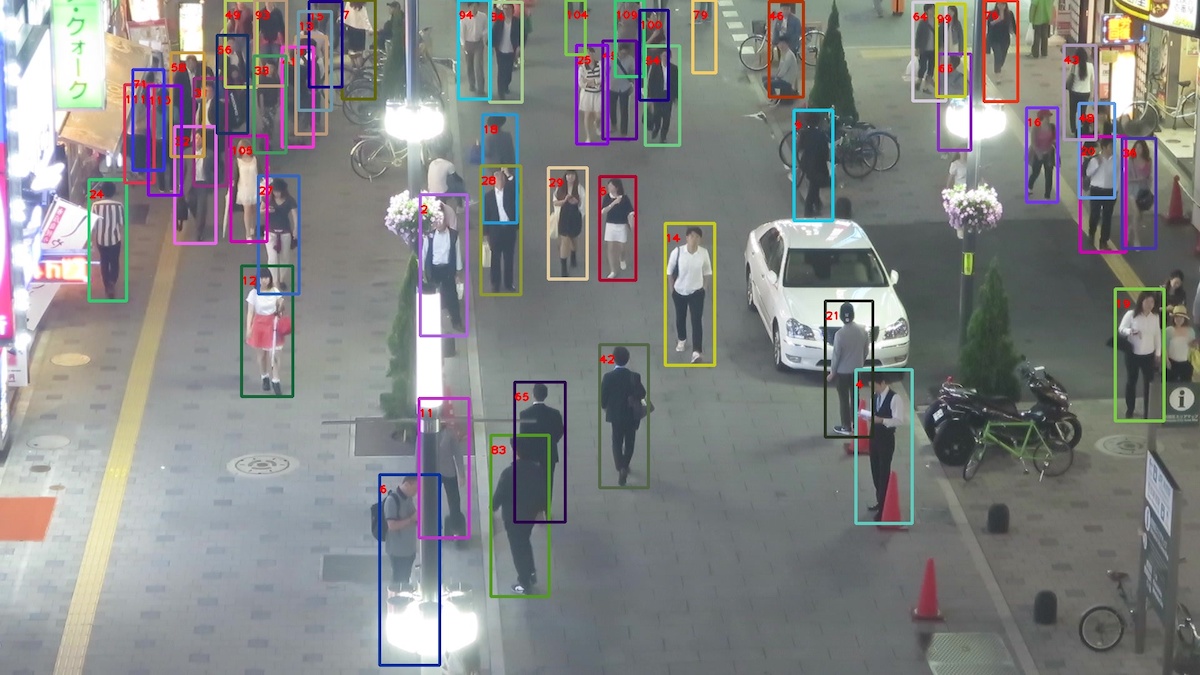} & \includegraphics[width=0.22\textwidth]{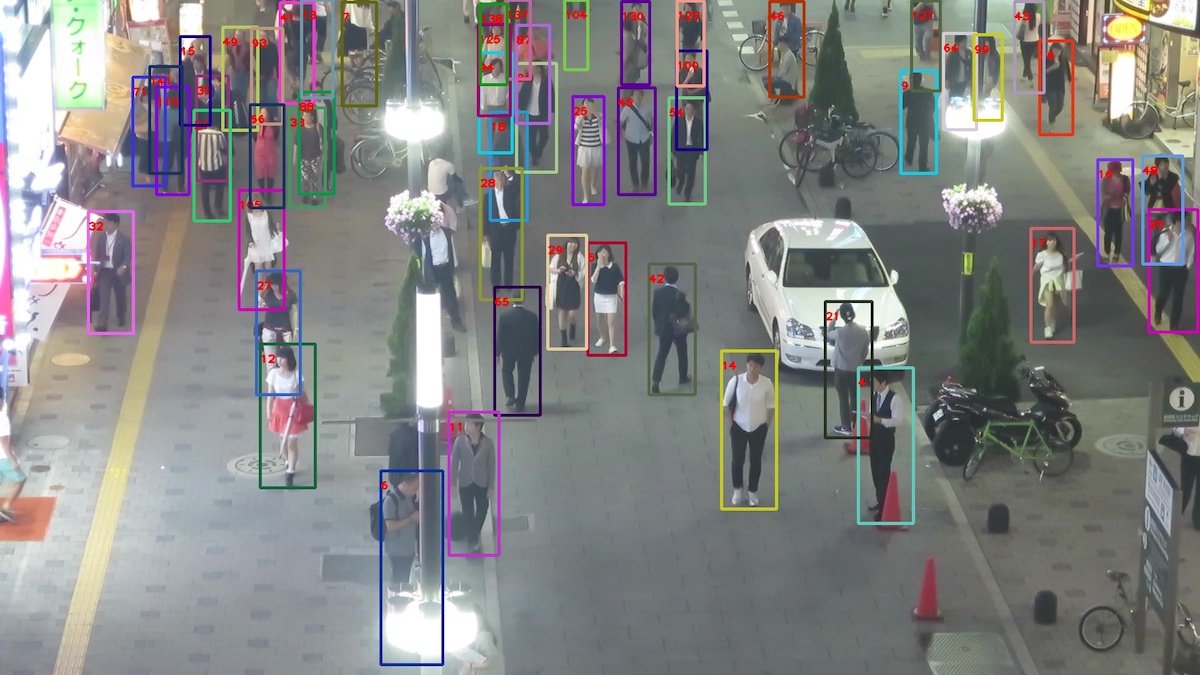} \\
        \end{tabular}
    \caption{These are some qualitative tracking results by our TDT-tracker on the MOT17-03~\cite{MOT16} sequence. The people in the video are clear and similar to the samples~\cite{wei2018person} that were used to train our teacher embedder. TDT-tracker tracks well on most of the people. For example, the person in white shirt with ID 20 (on the top right corner in Frame 1) is tracked successfully across the frames, although there is always a group of people around the person.}
    \label{fig:mot17-03}
\end{figure*}

\begin{figure*}[h]
    \centering
        \begin{tabular}{cccc}
        Frame 1 & frame 101 & frame 201 & frame 301 \\
        \midrule
        \includegraphics[width=0.22\textwidth]{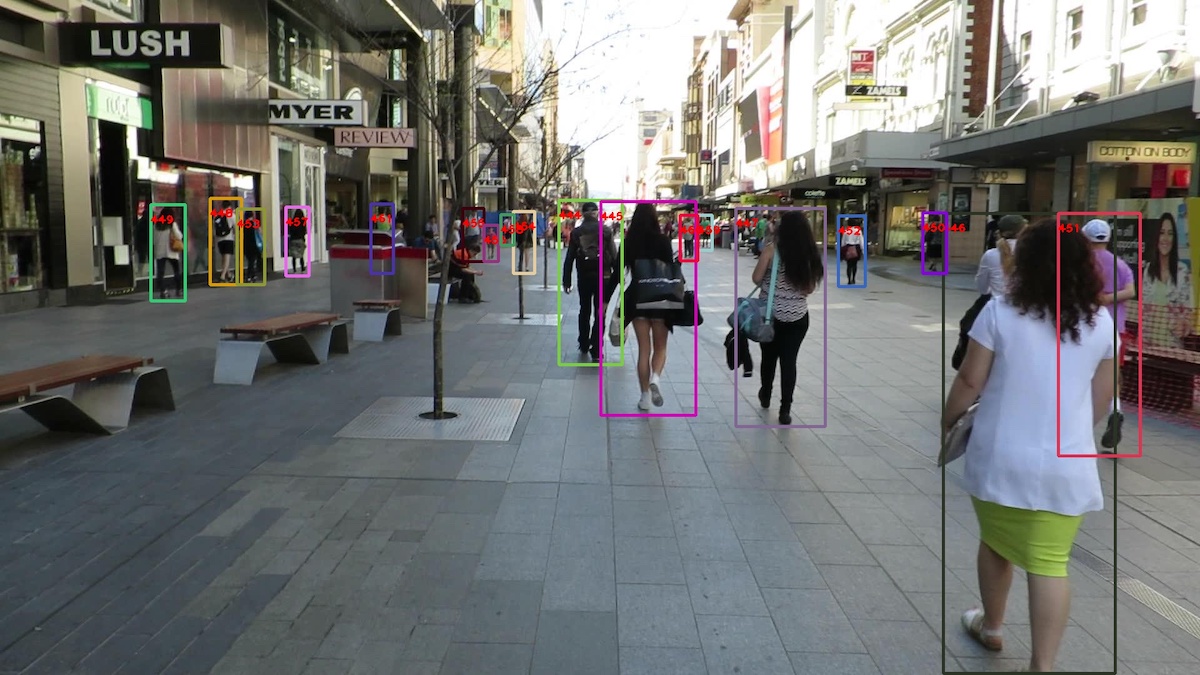} & \includegraphics[width=0.22\textwidth]{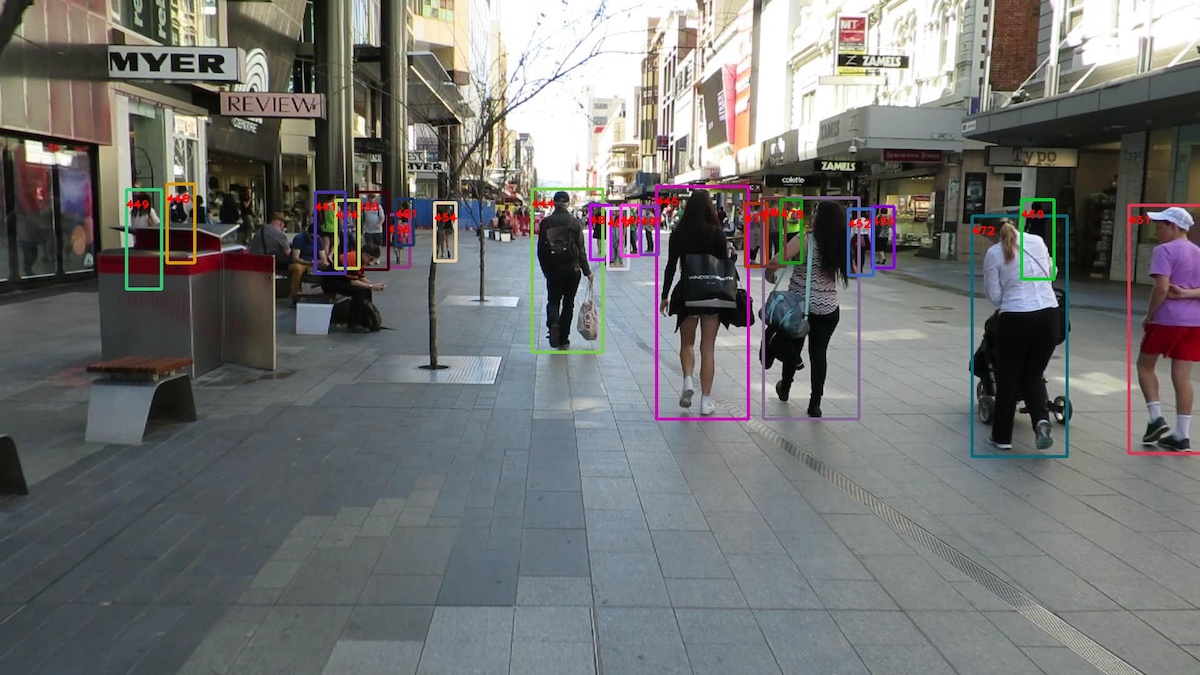} & \includegraphics[width=0.22\textwidth]{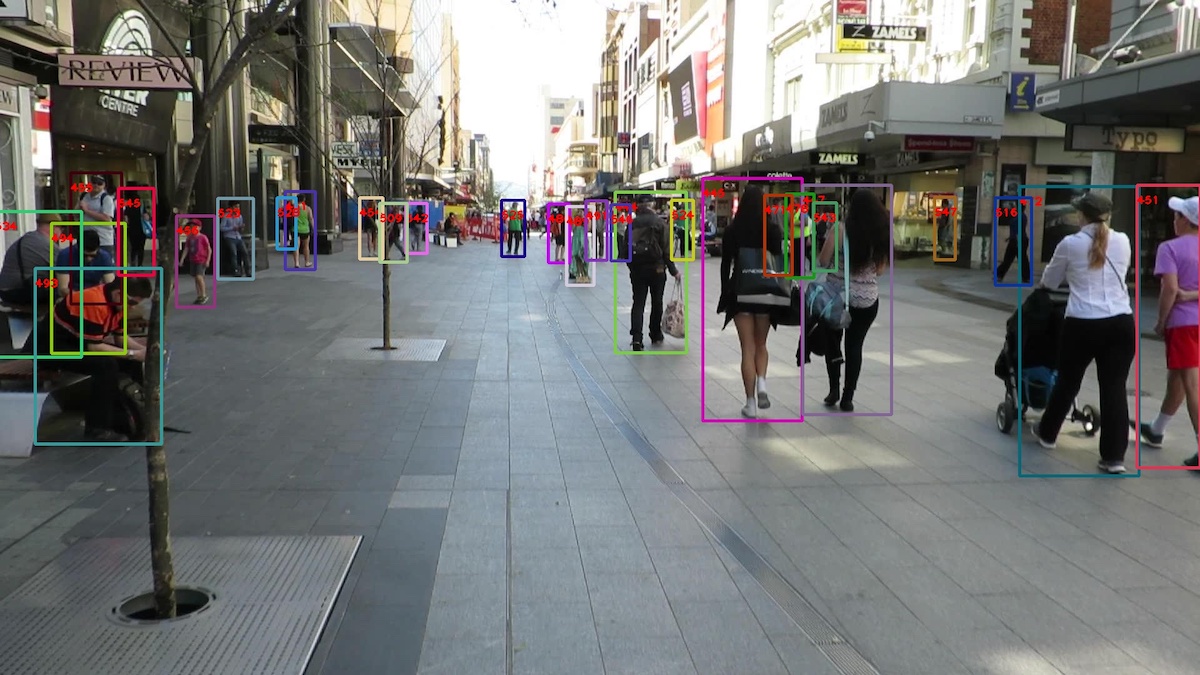} & \includegraphics[width=0.22\textwidth]{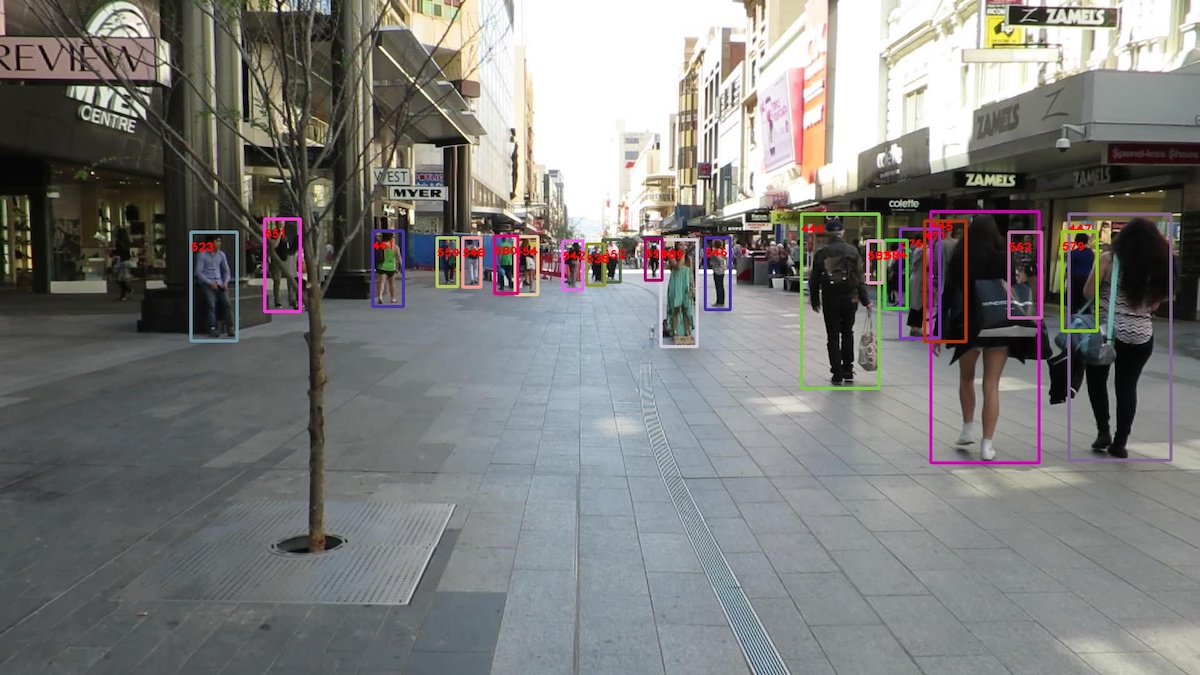} \\
        \end{tabular}
    \caption{These are some qualitative tracking results by our TDT-tracker on the MOT17-07~\cite{MOT16} sequence. These examples are used to show that our TDT-tracker can successfully track people who are occluded during the video. This attributes to the occluded samples in the training datasets~\cite{wei2018person} of our teacher embedder. For example, the person in green shirt with ID 461 (the fifth person from the left in Frame 1.) is successfully tracked across the frames although the person is partially blocked from the starting frame. Another example is the person in white clothes with ID 449 (the first from left in Frame 1). The lower body of the person is blocked in Frame 101 but the person is still successfully tracked.}
    \label{fig:mot17-07}
\end{figure*}

\begin{figure*}[h]
    \centering
        \begin{tabular}{cccc}
        Frame 1 & Frame 101 & frame 201 & frame 301 \\
        \midrule
        \includegraphics[width=0.22\textwidth]{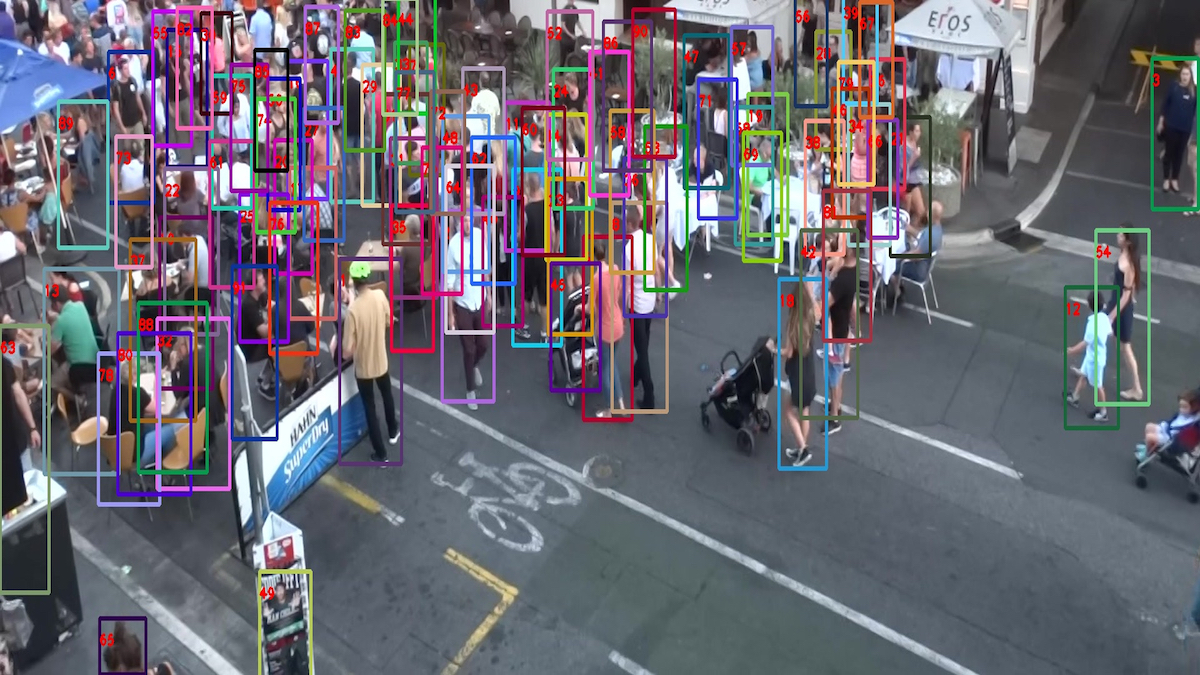} & \includegraphics[width=0.22\textwidth]{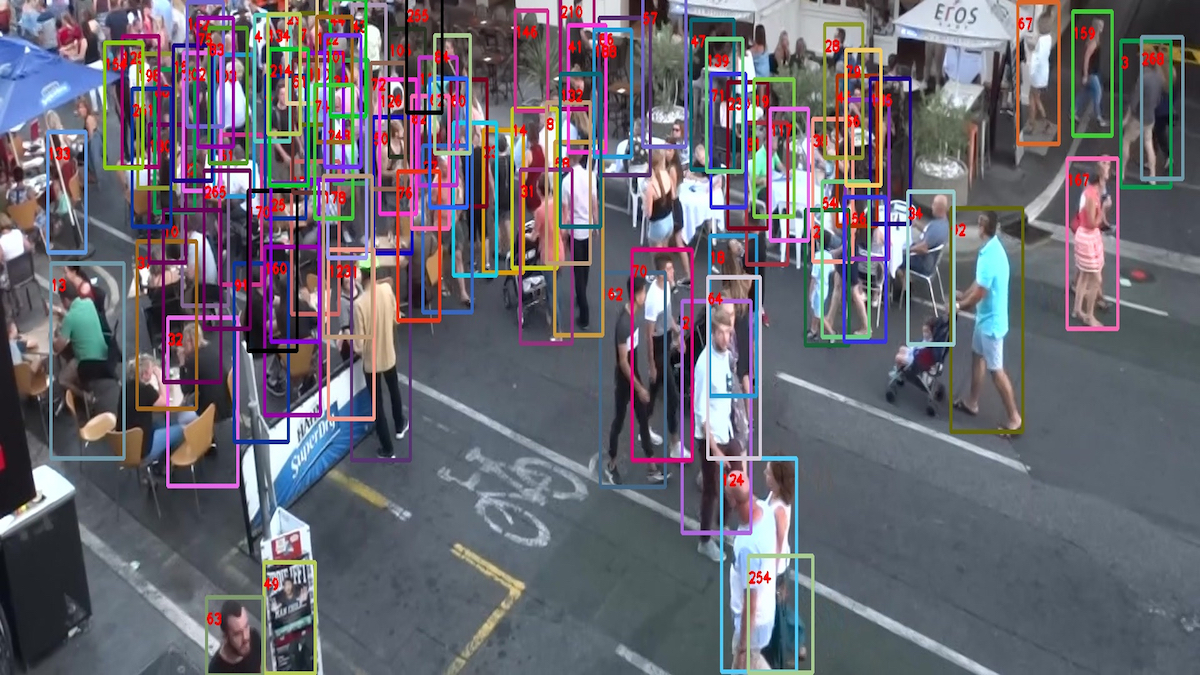} & \includegraphics[width=0.22\textwidth]{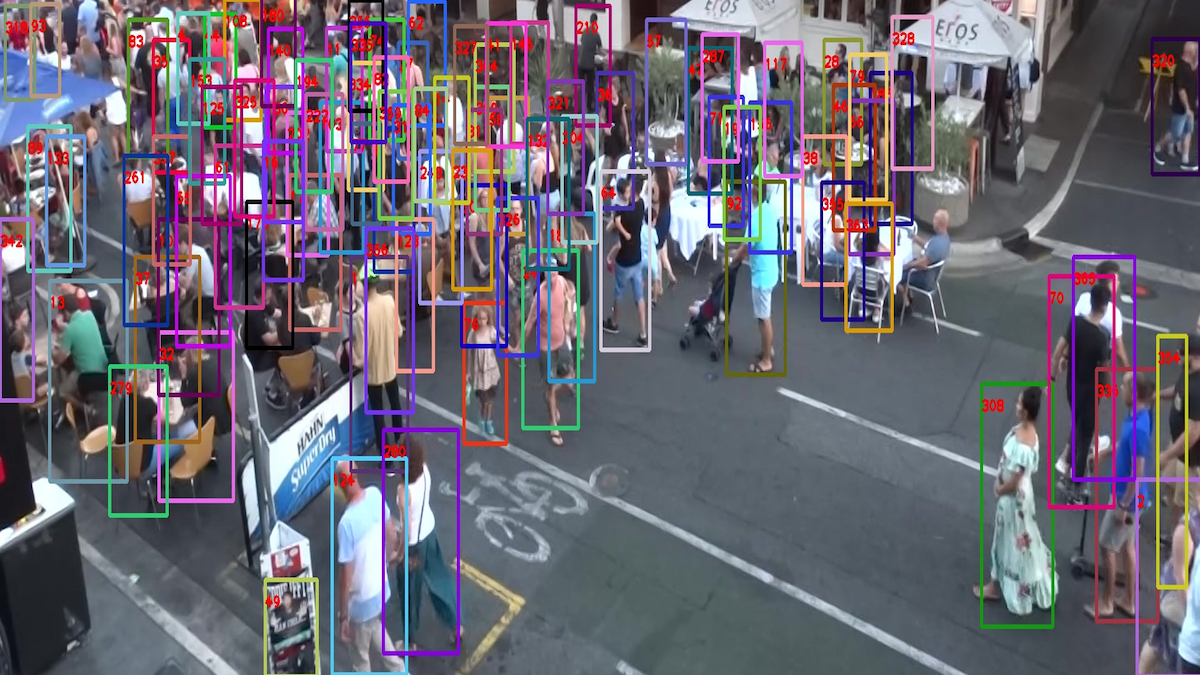} & \includegraphics[width=0.22\textwidth]{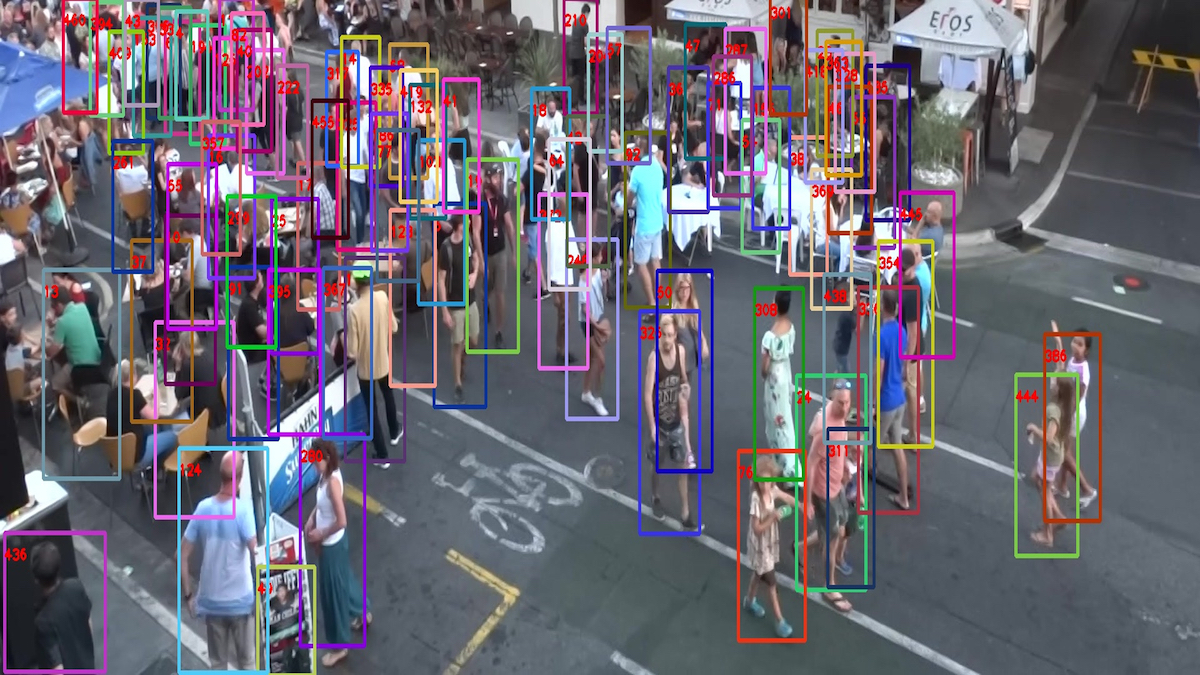} \\
        \end{tabular}
    \caption{These are some qualitative tracking results by our TDT-tracker on the MOT20-08~\cite{MOTChallenge20} sequence. Our TDT-tracker tracks many people poorly in the video. As we can see from the sampled frames, this is a very crowded scene with many overlapping among the people. The detector of TDT-tracker can detect most of the people accurately. The issue is likely that the embedding is not discriminative enough for tracking purpose. Our teacher embedder is not trained on people with this high level of overlapping. However, with increasingly better Re-ID networks as the teacher embedder, our TDT-tracker can improve without additional computational overhead.}
    \label{fig:mot20-08}
\end{figure*}

\Cref{fig:mot17-03}, \cref{fig:mot17-07}, and \cref{fig:mot20-08} show three qualitative examples of our TDT-tracker. The detailed analysis is in the caption of each figure.

% %%%%%%%%% REFERENCES
% {\small
% \bibliographystyle{ieee_fullname}
% \bibliography{egbib}
% }

\end{appendix}

\end{document}